\PassOptionsToPackage{numbers,compress}{natbib}
\documentclass{article}

\usepackage{wrapfig}
\usepackage{graphicx}
\usepackage{multirow} 
\usepackage{tabularx}
\usepackage{hyperref}
\hypersetup{
    colorlinks=true,
    citecolor={blue!50!black},
    linkcolor={blue!50!black},
    urlcolor={blue!80!black}
}
\usepackage{tcolorbox}
\tcbuselibrary{breakable}
\usepackage{float}
\usepackage{arydshln}

 \usepackage[main, final]{neurips_2026}
\usepackage[utf8]{inputenc} % allow utf-8 input
\usepackage[T1]{fontenc}    % use 8-bit T1 fonts
\usepackage{hyperref}       % hyperlinks
\usepackage{url}            % simple URL typesetting
\usepackage{booktabs}       % professional-quality tables
\usepackage{amsfonts}       % blackboard math symbols
\usepackage{nicefrac}       % compact symbols for 1/2, etc.
\usepackage{microtype}      % microtypography
\usepackage{xcolor}         % colors
\usepackage{amsmath}

\title{Masked Diffusion Language Models are Strong and Steerable Text-Based World Models for Agentic RL}

\author{%
  Darshan Deshpande \\
   \includegraphics[width=0.3cm]{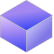} Patronus AI\\
  \texttt{darshan@patronus.ai} \\
}

\begin{document}

\maketitle

\begin{abstract}
Recent growth in different reinforcement learning (RL) techniques have surfaced a need for a wide variety of specialized training environments. These environments are typically hand-curated, with task and reward difficulties that are fixed rather than adaptive, making them ineffective training signals once a model's performance on the domain improves. As models continue to improve on these environments and reward signals grow increasingly sparse over longer horizons, the model encounters fewer diverse situations during rollouts, leaving it prone to overfitting on specific workflows or tool structures, also known as mode collapse. World models that simulate environment states have previously matched the performance of pure environment rollouts, making them a promising avenue for scaling diversity given that their outputs can be varied on-demand and at scale. However, autoregressive (AR) world models suffer from a fundamental left-to-right bias that prevents them from conditioning on globally interdependent state anchors such as tool schemas, prior turns, and expected outcomes. In this work, we (i) formalize text-based world modeling as a steerable transition-dynamics problem decomposed into initial environment state, task context, tool schemas, domain rules, and steering directives, and (ii) curate a dataset of 239,403 grounded state–action trajectories spanning nine open-source environments and twelve frontier model families. Using this dataset, we present a comparative study between AR LMs and masked diffusion language models (MDLMs), and show that MDLMs, by virtue of bidirectional anchor-aware denoising, produce better coherence, groundedness, and empirically validated rollout diversity than LLMs more than $4\times$ their total parameter size, with comparable inference latency. We introduce a plug-and-play GRPO training framework with deterministic state checks, and perform zero-shot transfer ablations on three out-of-distribution environments (ScienceWorld, ALFWorld, AppWorld) across three agent backbones from 1.2B–7B parameters (LFM2.5, Qwen3, Mistral), achieving absolute gains of up to 47\% over raw baselines without environment-specific fine-tuning. Finally, we conduct a behavioral analysis of failure modes under adversarial scenarios and a human evaluation centered on realism, outcome correctness, and training utility to showcase their reliability. We open source our work to encourage research in this direction~\footnote{Dataset: \url{https://huggingface.co/datasets/PatronusAI/world_model_corpus}\newline Training code: \url{https://github.com/patronus-ai/mdlm_world_modeling}}.
\end{abstract}

\section{Introduction}
% \section{Introduction}
% % Applications of RL / LLMs
% % Grip on Envs -> World Models
% % World Models
% % Research Questions
% % Answer summary for RQs
Reinforcement learning (RL) has transformed Large Language Models (LLMs) from passive sequence generators into decision-making agents capable of operating in complex and dynamic tool-use environments~\citep{shao2024deepseekmath_grpo}. Such agents have demonstrated strong performance across diverse domains including web navigation~\citep{hu-etal-2025-os}, code generation~\citep{dong2025survey}, and enterprise automation~\citep{chen2025sheetagent}. Unlike static reasoning tasks, agents deployed in multi-turn settings must maintain memory across dialogue rounds, perform sequential decision-making, and adapt to environmental feedback. Yet, training agents in multi-turn settings introduces compounding challenges: reward signal sparsity~\citep{cui2025processreinforcementimplicitrewards}, distributional shift across turns~\citep{xue2025simpletir}, and limited generalization beyond the training environment~\citep{xi2026can}. 

\begin{figure*}[t]
    \centering
    \includegraphics[width=\linewidth]{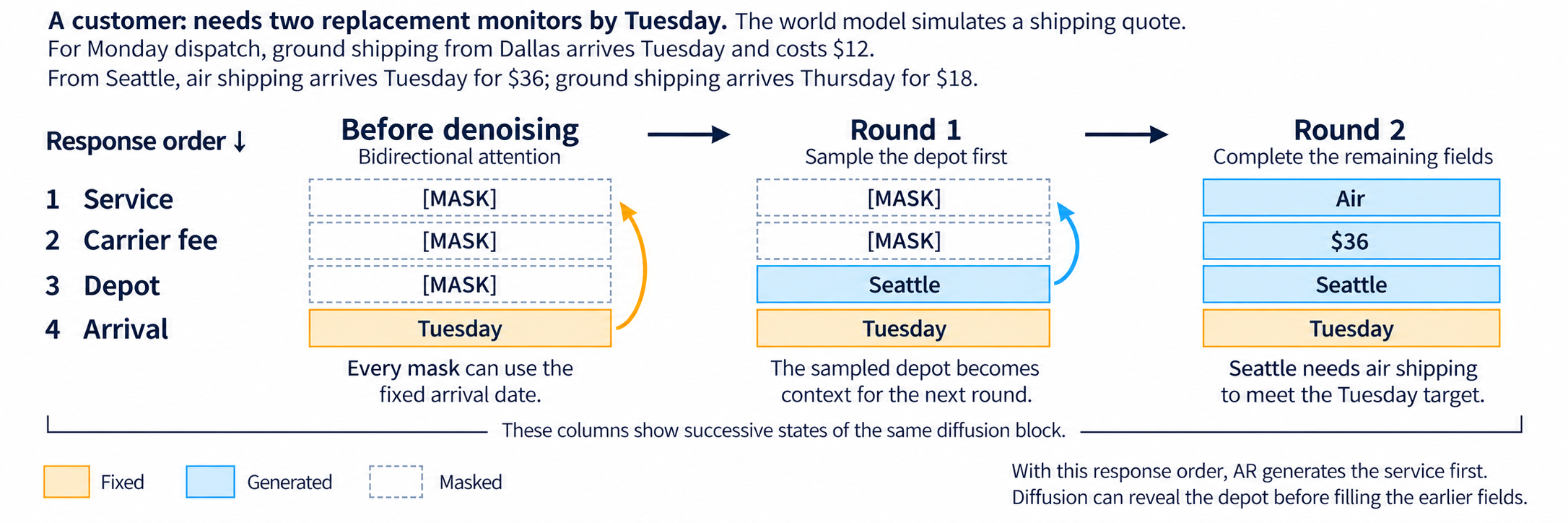}
\caption{Bidirectional attention conditions generation on the fixed arrival date. Iterative denoising reveals the depot first, then uses it to complete the earlier service and fee fields.}
    \label{fig:mdlm_vs_ar}
\end{figure*}

A fundamental bottleneck constraining agent training is the availability, diversity, and scalability of environments~\cite{chu2025sft, xi2025agentgym}. Standard RL environments are typically hand-engineered and static, causing agents to develop brittle policies that fail under distributional shift~\cite{padakandla2021survey, aryan2025abidegym}. Curriculum learning and unsupervised environment design have been proposed to address this~\cite{liang2024eurekaverse}, but constructing effective curricula still demands significant domain expertise. Domain randomization~\cite{mehta2020active} and procedural content generation~\cite{kim2025multi} have shown to improve inter-episode diversity, yet the environment remains fixed within each episode, leaving intra-episode adaptation largely unaddressed. Together, these limitations motivate learned environment simulators that can generate diverse, realistic dynamics at scale without hand-crafted engineering.

An alternative known as world models are generative models that predict future environment states conditioned on past states and actions. They offer a principled solution to this scalability and diversity problem~\cite{he2025pre}. Early works such as DreamerV3~\cite{hafner2023mastering} demonstrate that a world model can internalize environment dynamics and support imagined rollouts for planning, generalizing across over 150 diverse tasks with a single configuration. More recently, LLM-based text world models have emerged as a lightweight and scalable paradigm~\cite{li2025word}. \citet{li2025word} show that well-trained world models maintain coherent latent state, scale predictably with data and model size, and consistently improve downstream agent performance. Works such as~\citet{wu2025rlvr} further demonstrate that reinforcement learning from verifiable rewards can improve world model fidelity in both language and video settings, making them a promising direction. However, a critical limitation of autoregressive models (ARs) as textual world models is their left-to-right generation bias: environment states are globally interdependent, yet causal models are architecturally constrained to condition only on preceding context, making reliable steering harder and susceptible to hallucinations~\cite{alansari2026largelanguagemodelshallucination}. While techniques like chain-of-thought~\cite{wei2022chain} or Monte-Carlo-Tree-Search~\cite{xie2024monte} have been used for exploration and reduce hallucinations~\cite{dhuliawala2024chain}, such techniques increase inference latency, making them unsuitable for very long, batched agent rollouts for GRPO-like~\cite{shao2024deepseekmath_grpo} RL algorithms. 

% Diffusion models, progress in mdlms and why they are sutiable for WM
Masked diffusion language models (MDLMs)~\cite{nie2025largelanguagediffusionmodels, bie2025llada2, cheng2025sdar, liu2025wedlm} offer a structurally different alternative. By learning to iteratively denoise corrupted token sequences, MDLMs acquire bidirectional context over the full observation and naturally model global consistency which we show is essential for a faithful environment simulation. Unlike AR models, MDLMs are not subject to the causal masking constraint, allowing them to condition predicted states on both preceding and succeeding context within a state. Moreover, their denoising training objective encourages learning smooth, structured distributions over state spaces, which, in this paper, we hypothesize leads to more coherent and ecologically grounded environment rollouts. While diffusion models are standard for computer vision world models~\cite{valevski2024diffusion, bai2025masks, ding2025understanding}, their discrete, text counterparts, MDLMs have not been systematically evaluated as world models for textual environment simulations.

% Research Questions
In this work, we address the following research questions:

\begin{enumerate}
\item How do masked diffusion language world models compare against causal large language world models on in-domain and out-of-domain environment simulations?
\item Does training RL agents on MDLM-generated rollouts yield measurable downstream performance improvements on held-out environments compared to AR-rollout baselines?
\item Do MDLM produced environment rollouts showcase a high degree of realism, correctness and training utility according to human experts? What are some common failures of such world models?
\end{enumerate}

Through extensive experiments across different model sizes, we find that small MDLMs (8B parameters), including SDAR~\cite{cheng2025sdar}, LLaDA-2.1~\cite{bie2026llada2}, and WeDLM~\cite{liu2025wedlm}, consistently outperform finetuned causal models including Qwen-3.5~\cite{qwen35blog}, Nemotron-3~\cite{nvidia_nemotron_nano_v3_2025}, GLM-4.7-Flash~\cite{5team2025glm45agenticreasoningcoding}, and GPT-OSS-20B~\cite{openai2025gptoss120bgptoss20bmodel} at scales up to 35B mixture-of-expert parameters and 27B dense AR models. This result suggests that the causal inductive bias, rather than model capacity, is the primary bottleneck in LLM-based world modeling. While previous works such as~\citet{li2025word} require fine tuning of models for downstream applications, we show that GRPO training in a zero shot setting with MDLM generated trajectories leads to performance improvements of up to 47\% across LiquidAI~\cite{liquidai2025lfm2}, Qwen~\cite{yang2025qwen3} and Mistral~\cite{jiang2023mistral} families ranging from 1.2-7B parameters. Finally, our human evaluation studies also show that these models achieve high human alignment scores across the training utility, realism and outcome correctness metrics with the four annotator averages reaching 4.75, 4.25 and 4.50 on a 1-5 Likert scale with high inter-annotator alignment ($\alpha \geq$ 0.89). 

\section{Relevant Work}
\paragraph{LLMs as synthetic data simulators}
LLMs have recently been applied to large scale synthetic data generation tasks for post-training~\cite{nadas2503synthetic, long2024llms, alismail2025survey}. Common techniques for diversification and scalable generation cover persona based diversification~\cite{ge2024scaling, wang2025deeppersona}, document or web-grounded techniques~\cite{maini2024rephrasing, su2025nemotron, zhu2025real, lupidi2024source2synth}, workflow-based grounding~\cite{mitra2024agentinstruct, sudalairaj2024lab, Riaz_2025} and long horizon generation~\cite{he2025scaling, li2025wildlong, gao2025nextlong}. As RL techniques continue to mature and scale, several works have suggested techniques to scale tool simulations~\cite{goldie2025synthetic, liu2024apigen} and agent-human interplay for trajectory diversification~\cite{wu2025instruct, prabhakar2025apigen}. Since the release of the GRPO algorithm~\cite{shao2024deepseekmath_grpo}, post-training focus has shifted towards specialized training environments with binary verifiable rewards~\cite{wen2025reinforcement, wu2025rlvr, da2025agent}. The field has since seen a rapid surge of diverse training and evaluation environments covering coding~\cite{jimenez2023swe, merrill2026terminal}, computer use~\cite{OSWorld, rawles2024androidworlddynamicbenchmarkingenvironment, zhou2023webarena, wei2025browsecompsimplechallengingbenchmark}, customer service\cite{barres2025tau2} and agentic safety~\cite{debenedetti2024agentdojo}. Works such as~\citet{song2026envscaler} have covered scaling of environments but since task diversity in isolation is not enough and rewards and tooling need to scale with model performance~\cite{helff2026llmsgamingverifiersrlvr, chen2025scalingagentlearningexperience}, we have recently observed a rise of text-based world models~\cite{li2025word, wang2026agentworldmodelinfinity, chen2025scalingagentlearningexperience} that aim to improve training signal by helping diversify model experiences in a controlled manner. In this work, we attempt to further showcase abilities of models at simulating realistic environment behaviors to improve downstream RL performance with the hope to move towards RL in imaginative worlds, similar to~\citet{li2024open}. 

\paragraph{MDLMs as controllable models}
An ideal world model must be fast, high fidelity, generalizable and steerable~\cite{gao2024vista, genie3}. AR language models exhibit a strong left-to-right bias which restricts diversity and editability and the inference costs for such AR generation scales linearly with the sequence length, making these models difficult to setup for efficient and high performance inference~\cite{wu2023ar}. The MDLM training procedure enables parallel token generation and bidirectional infilling which further improve their applicability for arbitrary state prediction~\cite{sahoo2024simple, nie2025largelanguagediffusionmodels}. Furthermore, works such as ~\citet{nie2025largelanguagediffusionmodels, sahoo2024simpleeffectivemaskeddiffusion, shi2025simplifiedgeneralizedmaskeddiffusion} address the reversal curse and show potential for high quality generations. Since it is compute intensive to train MDLMs, recent works such as~\citet{fu2025efficientdlmautoregressivediffusionlanguage, gong2025scalingdiffusionlanguagemodels} have shown that conversion of large pre-trained AR models to MDLMs is not only possible but also performant. Building on top of these works, optimization techniques such as block diffusion~\cite{arriola2025block} and joint threshold for merging Token-to-Token editing with conventional Mask-to-Token decoding during inference~\cite{bie2026llada2} have been progressively studied and shown to be effective at scaling. Parallely, ~\citet{nie2025scalingmaskeddiffusionmodels} studied the scaling laws of such MDLMs and show that they are comparable to AR models which increasingly makes them suitable for world modeling applications. As MDLMs continue to scale across capabilities and domains~\cite{cheng2025sdar, liu2025wedlm}, their applicability to world modeling and environment state simulation is still understudied. We hope to address this gap through this work.

\paragraph{World Modeling}
Since domain specific RL requires highly specialized and robustly curated environments~\cite{Da2025AgentRLVRTS, Cheng2025RevisitingRL, Zhang2025TheLO}, the field has started exploring ways to improve experience-based learning~\cite{chen2025scalingagentlearningexperience} to reduce their dependencies on non-scalable and hard to maintain environments. While world modeling is well-studied in the computer vision domain, ~\citet{li2025word} were one of the first to exhaustively benchmark and ablate AR model performance for this task. Other parallel works have since explored diverse task and tooling simulations~\cite{lyu2026mockworldsrealskills, he2026resynautonomouslyscalingsynthetic} for scaling environment generations. \citet{wang2025llmsscalablegeneralpurposesimulators} applied augmentation of retrieved information to ground the content of their predictions whereas ~\citet{wang2026agentworldmodelinfinity} were the first to propose an end to end pipeline for task, tool and deterministic reward generation. But such a system has three drawbacks. First, the pipeline heavily relies on correctness of each agent created component which makes it susceptible towards cascading errors from previous components or poor design of environment components. Secondly, the downstream scalability of such a pipeline is not exhaustively studied which makes generalizability questionable. Lastly, steering for specific model behaviors is hard and infrastructural stability issues could potentially worsen if agent is not strong at coding or loses context. To address these challenges, we first add steering objectives to ensure grounding in real tool schemas, expected  behaviors and desired outcomes to provide finer grained guidance. Secondly, we propose a plug-n-play framework for world model-based training by utilizing deterministic state checks which makes observability easier. Finally, we show generalization on six tasks spanning enterprise tool use tasks, coding environments, open ended APIs and text-based games.

\section{Masked Diffusion Language Models for Steerable World Modeling}
\label{sec:mdlm_for_wm}

\paragraph{Formalizing the World Model Objective.}
Extending the definition of~\citet{li2025word}, we formalize a world model as 
a parametric approximation of the environment's transition dynamics. For any textual environment, given a 
trajectory $\tau = (\{s_0, a_0\}, \ldots, \{s_T, a_T\})$ with states 
$s_t \in \mathcal{S}$ and actions $a_t \in \mathcal{A}$, we decompose each 
state as $s_t = (e_t, c_t, \mathcal{T}, \mathcal{R}, \mathcal{D})$, where 
$e_0$ denotes the \textbf{initial environment state} (observable world configuration, 
e.g., files, databases, UI available at the beginning of the rollout), $c_t$ the \textbf{task context} (user goal and 
conversational history), $\mathcal{T}$ the \textbf{tool schemas} (available 
tools and their signatures), $\mathcal{R}$ the \textbf{domain rules} 
(constraints on valid transitions), and $\mathcal{D}$ the \textbf{steering 
directives} (high-level instructions and safety constraints). The world 
model then learns
\begin{equation}
    p_\theta\!\left(c_{t+1} \mid e_{0}, c_{\leq t}, a_{\leq t};\, 
    \mathcal{T}, \mathcal{R}, \mathcal{D}\right),
\end{equation}
where $\mathcal{T}$, $\mathcal{R}$, and $\mathcal{D}$ remain fixed across 
the trajectory. In text-based environments, each $s_t$ is serialized as a token sequence $\mathbf{x}^{(t)} = (x_1^{(t)}, \ldots, x_L^{(t)})$, so the world model learns a distribution over structured token sequences conditioned on the history. Note that the world model represents $e_{\leq t}$ internally rather than emitting it as output, encouraging internal state tracking while avoiding 
context pollution. The fidelity of this learned distribution determines the quality of imagined rollouts for downstream agent training, and grounding generated states in environment variables and steering directives yields a natural and adaptable learning curriculum for the agent.

\paragraph{Masked Diffusion Language Models for World Modeling.}
Masked diffusion language models~\cite{nie2025largelanguagediffusionmodels, bie2026llada2, cheng2025sdar, liu2025wedlm} learn to predict a full next-state token sequence by iteratively denoising masked tokens, conditioned on an unmasked context. Given a clean next-state sequence during training $\mathbf{x}^{(t+1)} = (x_1^{(t+1)}, \dots, x_L^{(t+1)})$, the forward process samples a diffusion time $\tau \sim \mathcal{U}(0, 1]$ and independently replaces each token with a mask token $[\mathrm{M}]$ with probability $\tau$, yielding the corrupted sequence $\mathbf{x}^{(t+1)}_\tau$. A mask predictor $p_\theta$ is trained to recover the masked tokens conditioned on the history of past states $\mathbf{x}^{(\leq t)}$ and actions $\mathbf{a}^{(\leq t)}$. The training objective is:
\begin{equation}
    \mathcal{L}_{\text{MDLM}}(\theta) = -\mathbb{E}_{\tau,\, \mathbf{x}^{(t+1)},\, \mathbf{x}^{(t+1)}_\tau} \left[ \frac{1}{\tau} \sum_{k=1}^{L} \mathbf{1}\!\left[x_{k,\tau}^{(t+1)} = [\mathrm{M}]\right] \log p_\theta\!\left(x_k^{(t+1)} \,\Big|\, \mathbf{x}^{(t+1)}_\tau,\, \mathbf{x}^{(\leq t)},\, \mathbf{a}^{(\leq t)}\right) \right],
\end{equation}
where the indicator selects positions that were masked at diffusion time $\tau$. This objective is an upper bound on the negative log-likelihood of $\mathbf{x}^{(t+1)}$ under the model~\cite{nie2025largelanguagediffusionmodels}.

\paragraph{Steering via Conditional Masked Generation.}
Standard steering for LLMs is achieved by techniques such as speculative decoding~\cite{leviathan2022fast} or prefix conditioning~\cite{saito2023prefix}. While this is effective for open-ended steering, a tokenized environment state has symmetric dependencies among its fields and entity attributes, tool schemas, and reward signals constrain one another without respecting any particular order. An AR factorization $\prod_i p(x_i \mid x_{<i})$ is forced to commit to one ordering, whereas the MDLM objective is equivalent to an any-order AR likelihood bound \cite{ou2024radd} and therefore learns every conditional direction, including conditioning on arbitrary known positions at inference time, from the same training signal. A further advantage of MDLMs is that predicting masked tokens in parallel from a shared context eliminates the within-state error accumulation that occurs when an AR model conditions later tokens on an already-hallucinated prefix \cite{ding2024diffusion}. This, in addition to selective mask filling as shown in~\autoref{fig:mdlm_vs_ar} strongly motivates the use of MDLMs for this task.

\paragraph{Diversity of MDLM Rollouts.}
Beyond coherence, the iterative denoising process in MDLMs induces greater structural diversity in rollout samples than causal sampling from an AR model. This effect is not due to temperature alone. It arises because MDLMs sample both \emph{what} token to generate at each position and \emph{which} position to commit to at each step. \citet{gong2025diffucoder} show that increasing the sampling temperature in a masked diffusion language model diversifies not only token choices but also generation order, a degree of freedom that AR decoders do not possess, since they must always generate the leftmost unfilled position. AR decoders are prone to \emph{prefix mode collapse}~\cite{berglund2024reversalcursellmstrained}, in which high-probability opening tokens are committed early and subsequently constrain all later decisions. By contrast, MDLMs can fill in the middle or end of a state before its beginning, which helps avoid concentrating samples around a narrow set of prefixes \cite{vicentino2026autoregressive}. For state prediction, this is important because the rollout distribution should capture the true grounding in plausible next states given $\left(\mathbf{x}^{(\le t)}, \mathbf{a}^{(\le t)}\right)$, rather than collapsing onto a mode induced by an arbitrary serialization order. We observe the empirical advantage of this in \autoref{sec:results}.

\vspace{-0.5em}

\section{Experimental Setup}
\label{sec:expt_setup}

\paragraph{Environment Selection and Dataset Curation.}
To study our research questions, we specially curate a total of 239,403 trajectories with tasks spanning software engineering (\textsc{SWE-Smith}~\citep{yang2025swesmith}, \textsc{Coderforge}~\citep{CoderForge2026}), research automation (\textsc{DeepSearchQA}~\citep{gupta2026deepsearchqa}, \textsc{OpenResearcher}~\citep{li2026openresearcher}), customer service (\textsc{TauBench}~\citep{barres2025tau2}), and general-purpose tool use (\textsc{Gorilla/BFCLv4}~\citep{patil2025bfcl}, \textsc{Toolathlon}~\citep{li2025toolathlon}, \textsc{Pandora}~\footnote{\url{https://huggingface.co/datasets/danilopeixoto/pandora-tool-calling}}, \textsc{Webshop}~\citep{yao2022webshop}). Following the findings of~\citet{li2025word}, we generate these trajectories with the following set of models to encourage diversity: \textsc{Qwen-3.5-4B}, \textsc{Qwen-3.5-9B}, \textsc{Qwen3.5-397B-A17B-FP8}, \textsc{GPT-5.4-mini}, \textsc{GPT-5.4}, \textsc{GPT-OSS-120B}, \textsc{Claude-4.6-Sonnet}, \textsc{Claude-4.5-Haiku}, \textsc{Gemini-3-Pro}, \textsc{MiniMax-M2.1}, \textsc{GLM-5}, and \textsc{DeepSeek-V3.2}\footnote{All open-source model trajectories were generated using the Fireworks AI API. All closed models were queried via their respective official API endpoints.}. To evaluate fairly on out-of-domain environments, we select the fully deterministic SQL-only split from \textsc{Intercode}~\citep{yang2023intercode}, along with \textsc{OccuBench}~\citep{hu2026occubench} and \textsc{API Bank}~\citep{li2023api}, to assess industry-relevant specialized tools, tool schemas, and a wide range of general tool-use capabilities, respectively. After filtering, post-processing, and length diversification, the final training dataset contains \textbf{239,403} unique trajectory instances in the training split, with \textbf{19,454} validation trajectories and \textbf{20,847} held-out, stratified in-domain test trajectories reserved for evaluation. We provide full details of our dataset curation pipeline including trajectory generation, length diversification, environment state grounding, and dataset statistics in Appendix~\ref{app:dataset_curation}. To show downstream effectiveness of MDLMs as world models, we use \textsc{ScienceWorld}~\citep{wang2022scienceworld}, \textsc{ALFWorld}~\citep{ALFWorld20} and \textsc{Appworld}~\citep{appworld-acl24} as completely out of distribution environments covering entirely new tool schemas and open ended game navigation and action tasks that were excluded from the training set. More details about the environment setup and dataset distribution are available in Appendix~\ref{app:rl-dataset-curation} and \autoref{fig:data_split_distribution}.

\paragraph{Baselines.}
We compare against a suite of state-of-the-art open-source autoregressive (AR) LLMs spanning 8B–35B parameters, covering both dense and mixture-of-expert (MoE) configurations: \textsc{Qwen3.5-27B}~\citep{qwen35blog}, \textsc{GPT-OSS-20B}~\citep{openai2025gptoss120bgptoss20bmodel}, \textsc{GLM-4.7-Flash}~\citep{5team2025glm45agenticreasoningcoding}, \textsc{Nemotron-3-Nano-30B-A3B}~\citep{nvidia_nemotron_nano_v3_2025}, and \textsc{Qwen3.5-35B-A3B}~\citep{qwen35blog}. For MDLMs, we select the three strongest publicly available block-diffusion checkpoints: \textsc{WeDLM-8B}~\citep{liu2025wedlm} and \textsc{SDAR-8B/30B-A3B}~\citep{cheng2025sdar} (dense), and \textsc{LLaDA-2.1-mini (16B-1AB, MoE)}~\citep{bie2026llada2}. Since \textsc{SDAR-8B} and \textsc{WeDLM-8B} adapt the \textsc{Qwen3-8B}~\citep{yang2025qwen3} weights and change its objective to masked diffusion, we additionally include \textsc{Qwen3-8B} as a baseline to isolate the contribution of the diffusion training and inference procedure. For each model we report zero-shot, 3-shot, and fine-tuned scores to separate the effect of in-context conditioning from supervised adaptation.

\paragraph{Training.}
All models are fine-tuned with AdamW ($\beta_1=0.9$, $\beta_2=0.95/0.99$), gradient clipping (\texttt{clip=1.0}), and a cosine schedule with warmup, at a maximum sequence length of 16{,}384 tokens. We perform automated sweeps over learning rates $\{1\times 10^{-6},\, 5\times 10^{-6},\, 1\times 10^{-5},\, 5\times 10^{-5}\}$ and batch sizes $\{1, 2, 4, 8\}$ and report only the best scores. For MDLMs, we retain the block size used during the model's original instruction tuning to preserve coherence as per~\citep{cheng2025sdar}. Each experiment is repeated with three random seeds $\{42, 7000, 9000\}$, and we report the mean in~\autoref{tab:main_results} and ~\autoref{tab:diversity_results_table}. All training runs use ms-swift~\citep{zhao2024swiftascalablelightweightinfrastructure} for LLM training, dFactory~\footnote{\url{https://github.com/inclusionAI/dFactory}} for \textsc{SDAR} and \textsc{LLaDA-2.1-mini} training and \textsc{WeDLM}'s official training code~\footnote{\url{https://github.com/tencent/WeDLM}} with MagiAttention~\citep{magiattention2025} for \textsc{WeDLM-8B}. Further details in~\autoref{harness_optimizations}.

\begin{table*}[t]
  \centering
  \caption{Generation quality of MDLMs and AR models on in-domain and out-of-domain test sets. The best zero-shot result in each column is \textbf{bolded}, and the best three-shot result is bolded in \textbf{\textcolor{red}{red}}. For 3-shot prompting, only the best AR and MDLM baselines are shown. Full 3-shot results are in Table~\ref{tab:three_shot_results}, and results for finetuned non-Transformer architectures are in Table~\ref{tab:non_standard_arch}.}
  \label{tab:main_results}
  \footnotesize
  \setlength{\tabcolsep}{3pt}
  \begin{tabular}{l ccc ccc ccc ccc}
    \toprule
    & \multicolumn{3}{c}{\textbf{In-domain}}
    & \multicolumn{3}{c}{\textbf{API-Bank}}
    & \multicolumn{3}{c}{\textbf{OccuBench}}
    & \multicolumn{3}{c}{\textbf{Intercode (SQL)}} \\
    \cmidrule(lr){2-4} \cmidrule(lr){5-7} \cmidrule(lr){8-10} \cmidrule(lr){11-13}
    \textbf{Model}
    & B-1 & R-L & MAUVE
    & B-1 & R-L & MAUVE
    & B-1 & R-L & MAUVE
    & B-1 & R-L & MAUVE \\
    \midrule
    \multicolumn{13}{l}{\textit{Baseline Models}} \\
    \midrule
    Qwen-3-8B          & .136 & .286 & .148 & .453 & .421 & .301 & .524 & .523 & .705 & .406 & .397 & .699  \\
    Qwen-3.5-27B       & .152 & .221 & .187 & .536 & .611 & .505 & .746 & .755 & .810 &.516 & .614 & .799 \\
    Qwen-3.5-35B-A3B   & .125 & .233 & .189 & .488 & .585 & .499 & .667 & .680 & .811 & .522 & .598 & .780 \\
    Nemotron-3-Nano-30B& .120 & .152 & .223 & .412 & .316 & .510 & .675 & .667 & .822 & .486 & .465 & .785 \\
    GLM-4.7-Flash      & .130 & .266 & .232 & .486 & .289 & .532 & .741 & .739 & .815 & .537 & .516 & .790 \\
    GPT-OSS-20B        & .125 & .191 & .231 & .436 & .384 & .400 &  .716 & .717 & .798 & .544 & .540 & .788 \\
    LLaDA-2.1-mini     & .146 & .158 & .168 & .401 & .444 & .452 & .648 & .655 & .811 & .412 & .424 & .701 \\
    WeDLM-8B & .139 & .157 & .241 & .488 & .522 & .405 & .670 & .748  & .821 & .491 & .555 & .781 \\
    SDAR-8B & .145  & .240 & .252 & .503 & .581 & .605 & .637 & .727 & .828 & .328 & .466 & .731 \\
    SDAR-30B-A3B & .121 & .295 & .261 & .677 & .649 & .697 & .709 & .775 & .831 & .373 & .482 & .749 \\
    \midrule
    \multicolumn{13}{l}{\textit{3-shot Baseline Models (best AR / MDLM; full results in Appendix~\ref{app-sec:additonal-results})}} \\
    \midrule
    GLM-4.7-Flash      & .401 & .565 & .629 & .686 & .689  & .707 & .746  & .742 & .838 & .612 & .597 & .822 \\
    SDAR-30B-A3B & .401 & .601 & .666 & .601 & .662 & .702 & .700 & .713 & .849 & .621 & .641 & .886\\
    \midrule
    \multicolumn{13}{l}{\textit{Finetuned Autoregressive Models}} \\
    \midrule
    Qwen-3-8B          & .524 & .572 & .699 &  .712 & .781 & \textbf{.840} & .588 & .590 & .897 & .610 & .677 & .884 \\
    Qwen-3.5-27B        & .601 & .732 & .804 & .700 & .702 & .765 & .678 & .725 & .911 & .659 & .622 & .917 \\
    GPT-OSS-20B    & .462 & .461 & .810 & .697 & .730 & .762 & .669 & .730 & .942 & .643 & .695 & .897 \\
    GLM-4.7-Flash      & .570 & .588 & .891 & .672 & .705 & .766 & .723 & .792 & .950 & .629 & .670 & .949 \\
    Nemotron-3-Nano-30B& .545 & .506 & .731 & .685 & .681 & .789 & .671 & .788 & .922 & .612 & .645 & .956 \\
    Qwen-3.5-35B-A3B    & .630 & .730 & .932 & .691 & .689 &.791 & .766 & .772 & .960 & .646 & .653 & .956 \\
    \midrule
    \multicolumn{13}{l}{\textit{Finetuned Diffusion Models}} \\
    \midrule
    LLaDA-2.1-mini     & .623 & .713 & .968 & .677 & .682 & .789 & .698 & .677 & .891 & .602 & .621  & .901  \\
    WeDLM-8B           & .728 & .758 & .963 & \textbf{.748} & \textbf{.785} & .838 & .715 & .788 & .971 & .726 & .711 & .921 \\
    SDAR-8B            & .805 & .813 & .982
                       & .701 & .682 & .772
                       & \textbf{.795} & \textbf{.797} & \textbf{.979}
                       & .751 & .741 & .955 \\
    SDAR-30B-A3B & \textbf{.848} & \textbf{.853} & \textbf{.992} &.711 & .728 & .831 & .712 & .782 & .971 & \textbf{.783} & \textbf{.792} & \textbf{.965} \\
    \midrule
    SDAR-30B-A3B \textit{(3 shot)} & \textbf{\textcolor{red}{.905}} & \textbf{\textcolor{red}{.910}} & \textbf{\textcolor{red}{.995}} & \textbf{\textcolor{red}{.951}} & \textbf{\textcolor{red}{.954}} & \textbf{\textcolor{red}{.900}} & \textbf{\textcolor{red}{.904}} & \textbf{\textcolor{red}{.862}} & \textbf{\textcolor{red}{.983}} & \textbf{\textcolor{red}{.831}} & \textbf{\textcolor{red}{.834}} & \textbf{\textcolor{red}{.970}} \\
    \bottomrule
  \end{tabular}
\end{table*}

\paragraph{Metrics.}
We evaluate world models along three orthogonal dimensions corresponding to our research questions: generation fidelity (RQ1), downstream training utility (RQ2), and semantic correctness as judged by humans (RQ3). For generation fidelity, we report BLEU-1 and ROUGE-L to measure n-gram and longest-common-subsequence overlap with reference next-states, providing surface-form comparisons across the full suite of in-domain and out-of-domain test sets. We additionally report MAUVE~\cite{pillutla2021mauve}, which measures distributional alignment between generated and reference states in embedding space and is less sensitive to surface variation than $n$-gram metrics. This setup directly tests the central prediction of our anchor-aware analysis (\autoref{fig:mdlm_vs_ar}): if causal models produce prefix-consistent but globally incoherent completions while MDLMs condition bidirectionally on anchors, then MDLMs should exhibit stronger distributional alignment with reference states even when $n$-gram overlap is comparable. We deliberately avoid exact-match accuracy because tool responses can produce valid versions of the same underlying state, making exact match a weak metric. We acknowledge that no surface-form metric directly verifies semantic equivalence and hence, we complement these scores with diversity metrics Self-BLEU~\cite{montahaei2019jointlymeasuringdiversityquality} and Distinct-N~\cite{li2015diversity} to test the mode-collapse claim in~\autoref{sec:mdlm_for_wm}.
To measure downstream effectiveness of world models on held out environments, we use task success rate metric as proposed by each environment's goal state. Finally, we structure an independent human evaluation for studying the realism, training utility and objective correctness of world model predictions which we delineate in~\autoref{app_sec:human_eval}.

\section{Results and Discussion}
\label{sec:results}

\paragraph{RQ1: How do MDLMs compare against AR LMs for World Modeling Tasks?}
Table~\ref{tab:main_results} reports generation fidelity across in- and out-of-domain evaluation suites. We observe that in the zero-shot setting, MDLMs of comparable size outperform their AR counterparts on the distributional alignment metric (MAUVE), with \textsc{SDAR-30B-A3B} achieving the best zero-shot MAUVE on three of four splits. The gap is most pronounced on \textsc{API-Bank} (.697 vs.\ .532 for \textsc{GLM-4.7-Flash}) and on the in-domain split (.261 vs.\ .232). We analyzed the outputs carefully to observe that in zero-shot settings, AR models tend to produce more verbosity which hurts distributional closeness to concise API responses. 
% Few shot
Since LLMs are good few shot learners~\cite{brown2020language}, we tested three-shot prompting which narrows but does not close this gap. We notice that few shot examples help LLMs better understand the level of verbosity required and hence, models such as \textsc{GLM-4.7-Flash} and \textsc{Nemotron-3-Nano-30B} end up outperforming or matching the \textsc{SDAR-30B-A3B} on API Bank like deterministic datasets. However, since the \textsc{SDAR} model builds on the representations of the base \textsc{Qwen3-30B-A3B} model, its ability to improve with in-context examples is complementary to the MDLM objective as seen for the \textsc{WeDLM} and \textsc{LLaDA-2.1-mini} models as well. After fine-tuning, we observe that MDLMs dominate across all four splits and all three metrics. \textsc{SDAR-8B}, despite being more than $4\times$ smaller than \textsc{Qwen3.5-35B-A3B}, surpasses every AR baseline on in-domain MAUVE (.982 vs.\ .932) and on \textsc{OccuBench} MAUVE (.979 vs.\ .960). On \textsc{Intercode}, \textsc{SDAR-8B} is competitive on MAUVE (.955 vs.\ .956 for the best AR baselines, \textsc{Nemotron-3-Nano-30B} and \textsc{Qwen3.5-35B-A3B}) while achieving substantially higher BLEU/ROUGE (.751/.741 vs.\ .646/.653).

Three-shot \textsc{SDAR-30B-A3B} further pushes in-domain MAUVE to .995. Importantly, the \textsc{Qwen3-8B} baseline allows us to isolate the diffusion contribution as \textsc{SDAR-8B} and \textsc{WeDLM-8B} outperform fine-tuned \textsc{Qwen3-8B} across the board (e.g., +.283 in-domain MAUVE for \textsc{SDAR-8B}), demonstrating that gains arise from the masked diffusion objective rather than from base-model capacity or training data. Since \textsc{WeDLM-8B} retains \textsc{Qwen3-8B}'s strictly causal attention mask yet still beats it on 11 of 12 split-metric pairs, bidirectional attention alone does not explain the gain, and fine-tuned non-Transformer baselines (RWKV-7B, Falcon-H1R-7B) underperform both (\autoref{tab:non_standard_arch}). To test anchor conditioning directly, we fix $k\in\{1,2,8,16\}$ future anchor tokens in the target state of 250 OOD samples (within a single block for a block-size-32 \textsc{SDAR-8B} checkpoint). \textsc{SDAR-8B}'s accuracy rises from .90 to .94 with $k$, while \textsc{Qwen3.5-27B} stays at .86 since it cannot condition on anchors beyond its decoding position (\autoref{tab:anchor}). Together, these results indicate that the causal inductive bias, rather than parameter count, is the primary bottleneck in LLM-based world modeling.

\textbf{Diversity of MDLM rollouts.} 
Table~\ref{tab:diversity_results_table} reports rollout-level diversity for the strongest fine-tuned models across the out-of-domain sets. \textsc{SDAR-8B} achieves the lowest Self-BLEU and the highest Distinct-N and MAUVE, confirming the hypothesis in~\autoref{sec:mdlm_for_wm} that MDLM samples cover more of the plausible next-state distribution rather than collapsing onto prefix-induced modes. This is consistent with the observation of~\cite{gong2025scalingdiffusionlanguagemodels} that MDLM temperature jointly controls token choice and generation order, providing an extra axis of stochasticity unavailable to AR decoders. On the RL workload ($N{=}4{,}567$), \textsc{WeDLM-8B} and \textsc{SDAR-8B} run at $0.68\times$ and $1.80\times$ the latency of non-thinking \textsc{Qwen3.5-35B-A3B}, whereas enabling AR thinking to recover comparable fidelity costs $3.75$-$4.87\times$ (\autoref{tab:thinking-ablation}).

\begin{figure*}[t]
    \centering
    \includegraphics[width=\linewidth]{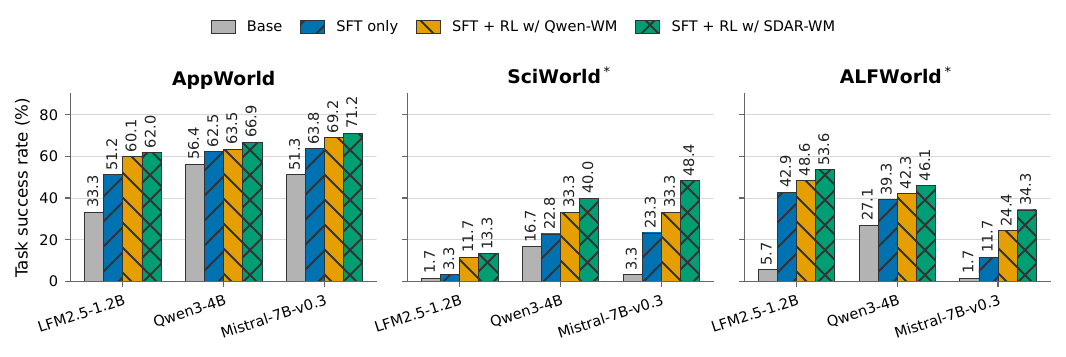}
    \caption{Task success rates across training methods on AppWorld, SciWorld, and ALFWorld for three policy models. \textsc{Qwen-WM} denotes \textsc{Qwen3.5-27B}, while \textsc{SDAR-WM} denotes \textsc{SDAR-8B}. Values above the bars report absolute success rates. Environments marked with $^*$ are made partially observable to evaluate long-context world-model behavior.}
    \label{downstream_results}
\end{figure*}

\textbf{RQ2: Does training agents on MDLM-generated rollouts improve downstream performance?}
\autoref{downstream_results} reports task success rates on \textsc{AppWorld}, \textsc{ScienceWorld}, and \textsc{ALFWorld}, across three agent backbones spanning 1.2B to 7B parameters. We select the \textsc{Qwen3.5-27B} and \textsc{SDAR-8B} due to their strong performance and ease of efficient dense model deployment (prompts in~\autoref{appsec:downstream_wm_prompts}). We also found that \textsc{Qwen3.5-27B} performance is equivalent to \textsc{Qwen3.5-35B-A3B} for these tasks in independent evaluations ($\pm0.5\%$ overall performance across two runs). Across all nine model-environment pairs, GRPO training with \textsc{SDAR}-derived world model rollouts (\textsc{SDAR-WM}) outperforms both SFT-only baselines and GRPO with Qwen-derived rollouts (\textsc{Qwen-WM}). The improvements are largest on partially observable, long-horizon environments like on \textsc{ALFWorld}, where \textsc{LFM2.5-1.2B} improves from $5.7\%$ (base) to 53.6\% with \textsc{SDAR-WM} ($+47.9$ absolute), and \textsc{Mistral-7B-v0.3} improves from $1.7\%$ to $34.3\%$ ($+32.6$ absolute). On \textsc{ScienceWorld}, \textsc{Mistral-7B-v0.3} jumps from $3.3\%$ to 48.4\% (+45.1 absolute), surpassing the \textsc{Qwen-WM} variant by 15.1 points. The consistent gap between \textsc{Qwen-WM} and \textsc{SDAR-WM} (averaging $+5.3$ points across all configurations) cannot be attributed to capacity, since both world models are fine-tuned on identical data but it instead reflects the higher fidelity and diversity of MDLM rollouts established in RQ1. Notably, these improvements occur in a zero-shot setting with respect to the held-out environments, contrasting with prior work from~\citet{li2025word} that required environment-specific fine-tuning and did not cover any environment formalizations. Although \autoref{downstream_results} reports the best configuration per sweep, \textsc{SDAR-WM} beats \textsc{Qwen-WM} in all 9 cells (sign test $p{\approx}0.004$) and in 31 of 33 hyperparameter-matched runs, with small checkpoint-to-checkpoint spread (e.g., $53.0\pm1.2$ vs.\ $48.1\pm1.4$ on \textsc{ALFWorld} with \textsc{LFM2.5}; \autoref{app:additional_rqs}).

\textbf{Why do MDLM rollouts help?} A post-hoc analysis of the logged GRPO rollouts links the RQ1 properties to these gains. Groups with zero reward variance, which contribute no gradient, fall from 47.3\% under \textsc{Qwen-WM} to 34.6\% under \textsc{SDAR-WM}. Within a group, the Distinct-2 diversity of predicted next states correlates with reward variance (Spearman $\rho{=}0.44$, 31k groups), and across the nine cells the difference in informative-group fraction predicts the success gap ($\rho{=}0.80$, $p{\approx}0.01$). \textsc{SDAR-WM} also references entities absent from the environment state less often (3.9\% vs.\ 8.7\% of turns), though it is worse on repeated pagination (3.8\% vs.\ 1.1\%) and unique-ID corruption (2.9\% vs.\ 0.6\%). These results are correlational but consistent with gains arising from more informative GRPO groups and fewer ungrounded states.

\begin{table*}
    \centering
    \caption{Aggregated Self-BLEU, Distinct-N (N=$\{1,2,3,4\}$) and MAUVE scores for fine tuned world models across test splits.}
    \label{tab:diversity_results_table}
    \begin{tabular}{lccc}
    \toprule
        Model & Self-BLEU ($\downarrow$) & Distinct-N ($\uparrow$) & MAUVE ($\uparrow$)\\
        \midrule
        \textsc{Qwen3.5-35B-A3B} & .690 & .253 & .889 \\
        \textsc{Qwen3.5-27B} & .670 & .321 & .864 \\
        \textsc{GPT-OSS-20B} & .659 & .228 & .867 \\\midrule
        \textsc{SDAR-8B} & \textbf{.601} & \textbf{.385} & \textbf{.902} \\
        \bottomrule
    \end{tabular}
\end{table*}

\textbf{RQ3: Human evaluation of realism, correctness, and training utility.}
We complement automatic metrics with a human study conducted by four industry experts recruited from Upwork\footnote{\url{https://upwork.com}}, each with at least two years of experience with LLM-based agentic harnesses. Annotators independently rated 100 \textsc{SDAR}-generated next states on a 1-5 Likert scale across three dimensions: \emph{realism}, \emph{outcome correctness}, and \emph{training utility} (rubric and guidelines in Appendix~\ref{app_sec:human_eval}). As we can observe in Table~\ref{tab:human_eval}, \textsc{SDAR} achieves means of 4.75 on realism, 4.25 on outcome correctness, and 4.50 on training utility, with Krippendorff's $\alpha$~\citep{krippendorff2011computing} above 0.89 on every dimension, indicating significant inter-annotator agreement. Annotators noted strong adherence to steering directives, particularly for adversarial scenarios such as forced tool failures, suggesting that bidirectional conditioning leads to reliable steerability without loss of realism. Common failure cases flagged by annotators included incorrect numeric key type coercions (string vs.\ integer), corrupted API keys, and a tendency for the model to collapse into repeated tool errors once any earlier turn returned an error, similar to~\citet{anil2024many}'s observations. Next, we study the finer grained behaviors in MDLMs and AR models. 

\textbf{Behavioral and steerability analysis.}
We probe \textsc{SDAR}'s behavioral patterns under a curated set of adversarial scenarios, contrasting them with AR baselines. The scenarios cover (i) reasoning over database states that lack a direct answer but contain adjacent information, (ii) cases that force the model to act under information insufficiency, (iii) verbose tool outputs known to challenge MDLMs, (iv) varying trajectory length constraints, and (v) infeasible tasks given the available toolset and (vi) complex navigation scenarios that require strong state recollections like pagination of tool outputs. We observe four consistent patterns. First, MDLMs are largely agnostic to element ordering in steering instructions, but degrade into repetitions without a repetition penalty when tools demand emitting the entire database state (e.g., \texttt{get\_all\_queries}). While AR models are stronger at producing coherent texts as a result of good instruction tuning, we observed that AR models share this limitation but they instead devolve into producing ungrounded and hallucinated text instead of repetitions. Second, MDLMs are capable of performing basic database joins in parameter space and maintain strong consistency with prior trajectory events, making them well-suited for state simulation. Third, on infeasible tasks the model continues to produce non-degenerate environment states rather than terminating early, which is desirable behavior for exploratory rollouts. Fourth, at high temperatures MDLMs occasionally fall into block-level repetition patterns. These can be post-processed away and diminish considerably with larger MDLMs. Out of these patterns, two limitations persist across our study: (1) MDLMs struggle at producing well-formed and consistent API keys which we believe is due to a combination of post-training safety alignment of the base Qwen models and repetition due to small block size during diffusion, and (2) they exhibit pagination drift when tools require enumerated multi-page outputs. While these can be handled by optimizing steering, we expect both to improve as MDLM scaling and tool-use centric post-training continues to mature. We provide a few failure example snippets in~\autoref{appsec:wm-failures} and \autoref{app:additional_rqs} for reference.
\vspace{-0.5em}

\begin{table}[t]
\centering
\small
\caption{Human evaluation results across three metrics. Mean is computed over annotator ratings (1-5 Likert scale) and Krippendorff's $\alpha$ reports inter-annotator agreement. Definitions for metrics are in Appendix~\ref{app_sec:human_eval}}
\label{tab:human_eval}
\begin{tabular}{lcc}
\toprule
\textbf{Metric} & \textbf{Mean} & \textbf{Krippendorff's $\alpha$} \\
\midrule
Realism              & 4.75 & .932 \\
Outcome Correctness  & 4.25 & .891 \\
Training Utility     & 4.50 & .901 \\
\bottomrule
\end{tabular}
\vspace{-0.5em}
\end{table}

\section{Conclusion}
\vspace{-0.5em}

In this work, we present the first systematic study of MDLMs as text-based world simulators for agentic RL.  By formalizing world modeling as a steerable transition-dynamics problem with five grounded components and curating a 239,403-trajectory dataset across nine environments and twelve frontier model families, we showed that MDLMs, by virtue of their bidirectional, anchor-aware denoising, produce more coherent, grounded, and diverse environment rollouts than autoregressive LLMs more than $4\times$ their size, at $0.68$-$1.80\times$ the inference latency of a non-thinking AR baseline. Our zero-shot transfer experiments on \textsc{ScienceWorld}, \textsc{ALFWorld}, and \textsc{AppWorld} further demonstrate that GRPO training with MDLM-generated rollouts yields absolute performance improvements of up to 47\% across agents spanning 1.2B–7B parameters, without any environment-specific fine-tuning. Human evaluation (Krippendorff's $\alpha$ > 0.89 across realism, outcome correctness, and training utility) supports the reliability of these rollouts. Despite these gains, we observe failure modes including pagination drift, block-level repetition under verbose tool outputs, and brittle handling of structured fields such as API keys that limit the immediate use of small MDLMs as drop-in environment substitutes. Both MDLM and AR world models also drift over long horizons (state-check pass rates fall from 96.8\% to 91.7\% and from 96.1\% to 89.4\%, respectively, between $t{\le}5$ and $t{>}25$), so we claim bounded drift rather than superior long-horizon coherence, and our evaluation is limited to turn-based text environments. We encourage future work on steering, robustness and reward hacking tendencies of these world models along with the potentially induced biases  in the downstream model. 

\bibliography{neurips_2026}
\bibliographystyle{plainnat}
% %%%%%%%%%%%%%%%%%%%%%%%%%%%%%%%%%%%%%%%%%%%%%%%%%%%%%%%%%%%%

\appendix

\section*{Appendix}
\section{Human Evaluations}
\label{app_sec:human_eval}
To verify that our gains come from the masked diffusion process itself rather than simply from access to more data, we conduct a thorough human evaluation of the MDLM world model's outputs. Specifically, we study three independent metrics: realism, outcome correctness and training utility. We ensure that no harmful or sensitive content is present during the human evaluation process and that the 100 samples are thoroughly author validated for safety. Below are the complete human evaluation guidelines:

\begin{table}[t]
    \centering
    \caption{Annotation rubrics for the three evaluation dimensions.}
    \label{tab:annotation-rubrics}
    \begin{tabular}{clp{9cm}}
        \toprule
        \textbf{Score} & \textbf{Label} & \textbf{Description} \\
        \midrule
        \multicolumn{3}{l}{\textit{Training utility}} \\
        \midrule
        5 & Excellent & Correct and consistent outcome, realistic values, i.e.\ ideal training signal \\
        4 & Good      & Minor value errors, but correct outcome type \\
        3 & Mediocre  & Correct outcome type, but wrong schema or implausible values \\
        2 & Poor      & Wrong outcome type or severely hallucinated structure \\
        1 & Useless   & Completely wrong, nonsensical, or would teach incorrect behaviours, including incomplete outputs \\
        \midrule
        \multicolumn{3}{l}{\textit{Realism}} \\
        \midrule
        5 & Indistinguishable      & Looks exactly like a real API response \\
        4 & Mostly realistic       & Minor inconsistencies (e.g.\ slightly wrong field names, but still valid) \\
        3 & Partially realistic    & Notable hallucinations like invented fields that don't fit the domain \\
        2 & Implausible            & Schema or values are implausible for this domain \\
        1 & Completely unrealistic & Random or nonsensical data \\
        \midrule
        \multicolumn{3}{l}{\textit{Outcome correctness}} \\
        \midrule
        5 & Exactly correct      & Correct outcome type and all key facts \\
        4 & Mostly correct       & Correct outcome type with minor factual errors \\
        3 & Partially correct    & Correct outcome type but significantly wrong facts \\
        2 & Wrong outcome        & Wrong outcome type (e.g.\ predicted success when an error was expected) \\
        1 & Completely incorrect & Entirely wrong \\
        \bottomrule
    \end{tabular}
\end{table}

\begin{tcolorbox}[title=Annotation Guidelines, colback=gray!5, colframe=gray!50, breakable]

\textbf{Annotator Requirements:}
Before beginning, please confirm you meet all three of the following requirements:
Technical background: You are familiar with tool-calling LLMs, including tool schemas and how an environment responds to a given action.
\newline

\textbf{Education and experience:} 

\begin{enumerate}
    \item You hold a CS degree and have experience both using and building with AI agents.
    \item Age: You are at least 18 years old.
\end{enumerate}

\textbf{Qualification Task (Complete First):}
Before accessing the full dataset, you must complete a short qualification task. This is used to evaluate annotation quality and ensure consistency across annotators.
The qualification task consists of 5 sample items. These will be provided to you by the project coordinator before you receive the full dataset.
Complete the qualification samples independently and without rushing. You must treat them exactly as you would the real task.
Do not use any kind of AI for completing the task. AI model capabilities are not reliable for this task. Submit your qualification responses in the "Qualifying Tasks" tab of your assigned Google Sheet (see Submission Instructions below).
The project coordinator will review your responses and confirm whether you are cleared to proceed.
\newline

\textbf{Evaluating Predicted Next States (100 conversations):}

What you'll see
Each item presents a JSON formatted conversation. At the end of every conversation, there is an “action”. You must rate the next state that is predicted after the action.
What you'll do
Rate the Predicted Next State on three separate dimensions. Each dimension is rated on a scale of 1–5. Assign scores independently. A response can score differently across dimensions. You must make your ratings independent of the expected output since tool outputs can produce different states (for example, sometimes a tool errors out, sometimes returns correct responses)
\newline

\textbf{Dimension 1: \textit{training\_utility}}
\newline

\textit{Would this prediction be useful for training a model? Think: "If I were an AI model and I saw this output from a tool I called, would the output be useful to me to proceed normally and correctly with the task?"
}\newline

\textbf{Dimension 2: \textit{realism}}
\newline

\textit{Does the predicted response look like a real API response for this domain?}\newline

\textbf{Dimension 3: \textit{outcome correctness}}\newline

\textit{Does the model correctly predict the outcome of the action?}

[See \autoref{tab:annotation-rubrics} for exact rubric breakdown for Likert scale.]
\newline

\textbf{General Guidance:}
Be consistent. Apply the same standards across all items.
Consider each dimension independently and don't let one score influence another.
Focus on validity of the tool output and its possibility to appear in a realistic situation
If something looks hallucinated but happens to be plausible for the domain, use your domain judgment to calibrate.

\end{tcolorbox}

As observed in~\autoref{tab:human_eval}, the mean scores across four annotators remains high at 4.75, 4.25 and 4.5 for realism, outcome correctness and training utility respectively. We utilize Krippendorff's alpha~\cite{krippendorff2011computing} to calculate interannotator agreement and reliability and we observe a high degree of consistency between the annotators. Common issues surfaced by annotators covered incorrect data type of numeric fields (e.g. string vs integer or float), corrupted API keys in tool calls or ungrounded and unrealistic websearch content. One instance flagged by two annotators was the MDLM's tendencies to collapse into repeated tool errors once any arbitrary turn returns an error.  
%%%%%%%%%%%%%%%%%%%%%%%%%%%%%%%%%%%%%%%%%%%%%%%%%%%%%%%%%%%%

\section{Fairness of Evaluation}
Since world model outputs are non-deterministic and API responses are generally in JSON format, the keys of such JSON mappings are seldom ordered which can lead to deflation of precision-based BLEU and ROUGE scores. To avoid this bias, we apply the following preprocessing steps to ensure fairness of evaluation. 

\begin{enumerate}
    \item We canonicalize JSON outputs to eliminate spurious mismatches: keys are sorted alphabetically in ascending order, all whitespace is stripped from both predicted and reference outputs, and stringified numeric entries are normalized.
    \item For \textsc{API-Bank}, the schema specifies \texttt{input}, \texttt{output}, \texttt{exception}, and \texttt{api\_name}. Since \texttt{input} is shared across models and does not reflect output quality, we omit it from evaluation.
    \item We use the GPU implementation of MAUVE with $k = \max\!\left(2,\, \operatorname{round}\!\left(\min(|p|, |q|) / 10\right)\right)$, following the auto-sizing recommendation from the original MAUVE paper. This ensures consistency with prior evaluations conducted under the implementation of~\citet{pillutla-etal:mauve:jmlr2023}. We use the default GPT-2 implementation proposed by the authors for this task.
\end{enumerate}

In addition to this, specifically for AR models, we find that higher temperatures lead to hallucination and more verbose outputs. To balance the verbosity-correctness ratio we evaluate across $\mathrm{temperature}=\{0.5,0.7,0.9\}$ and find that 0.5 achieves the best balance. With temperatures closer to 0, MAUVE scores drop by up to 0.2 points consistently which aligns with reduced variance in greedy decoding token selections. 

\section{World Model Training Dataset Curation Details}
\label{app:dataset_curation}
Since no existing world modeling datasets exist as of the writing of this paper, to help compare causal and masked diffusion language models, we attempt to create a broad coverage dataset of real environment rollouts, capturing all unique state-actions pairs. We describe our complete dataset curation, filtering and analysis procedure below.

\paragraph{Trajectory Generation.}

For every environment selected, we setup the original environment as designed by the authors and generate complete rollouts with a variety of models. To ensure diversity of trajectories and broad behavioral coverage, we sample a trajectory-generating model uniformly at random from the following set: Qwen3.5-4B, Qwen3.5-9B, Qwen3.5-397B-A17B-FP8, GPT-5.4-mini, GPT-5.4, GPT-OSS-120B, Claude-4.6-Sonnet, Claude-4.5-Haiku, Gemini-3-Pro, MiniMax-M2.1, GLM-5, and DeepSeek-V3.2.\footnote{All open-source model trajectories were generated using the Fireworks AI API. All closed models were queried via their respective official API endpoints.} Sampling across this diverse set of frontier models introduces natural variance in reasoning style, verbosity, tool-calling patterns, and error recovery behavior. We select the model with a starting seed of 42 for reproducibility.
\paragraph{Trajectory Length Diversification.}
Since steering of a world model relies heavily on the present state-action pair, trajectory lengths are important to diversify for natural rollout behaviors. To introduce trajectory length variance and prevent the model from overfitting to fixed-length contexts, we apply two complementary processing techniques.

\textit{Middle Truncation.} For conversational histories exceeding 16,384 tokens (approximately 2\% of the full dataset), we apply middle truncation to preserve realistic long-horizon workflows while respecting context length constraints. Specifically, we retain the original system prompt and user prompt alongside the final $N$ exchanges, where $N \sim \mathcal{U}(0, N_{\max})$ and
\begin{equation}
    N_{\max} = \left\lfloor \frac{T_{\max} - T_{\text{sys}} - T_{\text{user}}}{\bar{t}} \right\rfloor,
\end{equation}
where $T_{\max} = 16,384$ is the context length threshold, $T_{\text{sys}}$ and $T_{\text{user}}$ are the token counts of the system and user prompts respectively, and $\bar{t}$ is the mean token length per exchange computed over the training corpus. When $N = 0$, only the system and user prompts are retained, representing the most aggressively truncated case and ensuring the model learns to predict environment states from minimal conversational context. This is a practically important capability for world models operating under constrained inference budgets. The removed portion of text is replaced with the placeholder token \texttt{[TRUNCATED \{N\_CHARS\} characters...]}, which teaches the model to parse and process partially observable long trajectories that may require truncation at training or inference time.

\textit{Sub-trajectory Extraction.} For the remaining trajectories, we randomly select a tool call boundary and truncate the trajectory at that point, treating the subsequent environment state as the ground truth prediction target. All utterances following the selected tool call are removed. We apply this transformation with a random 15\% probability to prevent unnecessary truncation of the dataset which ensured better long horizon understanding in our experiments. This naturally exposes the model to a wide range of trajectory prefixes and reinforces robust prediction at varying horizon lengths.
\paragraph{Environment State Grounding.}
To enable better world model steering, we use Claude-4.6-Sonnet to synthesize additional ground truth environment context for each trajectory. This context includes extra database states, expected task reward structure, behavior patterns, and tool schemas and definitions. Concretely, we prompt Claude-4.6-Sonnet in high reasoning mode to analyze the full trajectory and generate instructional hindsight in the form of a structured description of what environmental information would have been sufficient to predict the final state, conditioned on the observed intermediate states. To promote realism and prevent overfitting to templated outputs, we deliberately include contextually adjacent but non-essential information in the synthesized context, such as neighboring database columns and irrelevant rows. This trains the model to selectively attend to relevant portions of large environment state dumps, reflecting real rollout conditions where custom prompt engineering is infeasible and ground truth database states are too large.
We apply instruction augmentation to approximately 80\% of the dataset to preserve the model's general approximative capacity and reduce overfitting to hindsight instructions. After filtering, post-processing, and truncation, unaugmented instances constitute 17.2\% of the final training dataset. To understand if the augmentations are necessary and do not produce unnecessary noise, we perform a minimalistic human evaluation. We select two independent annotators and task them to evaluate the trajectories in isolation and again with the grounding and steering objectives and noticed a high approval rate of 87\% for instances with grounding and steering instructions. Most disapproval cases flagged by the annotators belonged to deterministic prediction such as re-executing a terminal command without changes or formulaic tool calls. On the other hand, most agreements were seen in non-deterministic environments such as deep research or customer service tasks. The prompt used for this grounding is given below.

\begin{tcolorbox}[title=HINDSIGHT INSTRUCTION SYNTHESIS PROMPT, colback=gray!5, colframe=gray!50, breakable, label={hindsight_gen_prompt}]
\small
\begin{verbatim}
## ROLE AND OBJECTIVE
You are an expert environment annotator for training world-model agents on
tool-use trajectories. Given a trajectory state and the next agent action,
synthesize a "hindsight instruction": structured ground-truth environment
context, tool schemas, and behavioral rules that would have been sufficient
to predict the next environment observation, conditioned on the observed
intermediate states. These instructions teach the model to ground predictions
in environment mechanisms rather than templated heuristics, and to selectively
attend to relevant portions of large environment state dumps.

Reason carefully and silently before writing the final instruction.

## DOMAIN TAXONOMY
Decide which of the following four domains best matches the trajectory, then
apply the corresponding section template. If ambiguous, default to "Tool Use".

- SWE                : code/repo work, terminal/editor tools, tests, builds.
                       Examples: str_replace_editor, bash, pytest contexts.
- Tool Use           : (near-)stateless single-tool API calls.
                       Examples: calculator, calculate_age, format converters,
                       weather lookups with no persistent backing store.
- Customer Service   : multi-turn agent operating under a written policy with
                       user/account/profile state and back-office tools.
                       Examples: telecom support, retail returns, banking.
- Deep Research      : open-web research with search and browsing tools; the
                       environment is the unbounded web rather than a fixed DB.
                       Examples: web_search + read_webpage trajectories.

## INTERNAL REASONING CHECKLIST
Before writing, work through:
1. Domain classification (one of the four above).
2. Which environment(s) and toolset(s) are active.
3. What state is observable from the conversation; what is implied but hidden.
4. Which tool the next action invokes; expected parameters and return shape.
5. The causal mechanism that determines the response (DB lookup, parameter
   validation, file slice, web fetch, policy gating).
6. Distractor / non-essential context to inject (irrelevant rows, sibling
   tools, unused credentials, neighboring file paths).
7. Minimal context that closes the prediction gap.

## INPUT
You will receive a record with the following fields:
- `state`     : full conversation up to this point: system prompt, tool
                catalog, prior assistant/tool turns.
- `action`    : the next agent action (one or more tool calls).
- `outcome`   : (optional) the actual environment/tool response, used only
                to determine relevance, never quoted verbatim.

## SECTION INVENTORY (canonical order)
Include sections in this exact order, using Markdown level-2 headers (`##`).
Never include preamble or postscript.

  1. ## ENVIRONMENT STATE        - CONDITIONAL
  2. ## TASK CONTEXT             - REQUIRED
  3. ## TOOL SCHEMAS             - REQUIRED
  4. ## DOMAIN RULES             - CONDITIONAL
  5. ## STEERING DIRECTIVES      - CONDITIONAL
  6. PREDICTION TARGET: <...>    - REQUIRED (single final line, no `##`)

Conditional sections are included only when informative. The PREDICTION TARGET
line is ALWAYS the final line of the output.

## CONDITIONAL-SECTION INCLUSION RULES

ENVIRONMENT STATE
- INCLUDE when stateful backing data exists: DB rows, files, customer
  profiles, repo state, prior tool returns, session/auth flags.
- OMIT for Deep Research (open web; no fixed state).
- For Tool Use: include only the input record / derived state, kept short.

DOMAIN RULES
- INCLUDE when there are policies, conventions, mechanisms, or constraints
  beyond the bare schema: telecom policy, repo conventions, research
  methodology, account/line/bill semantics, validation rules.
- OMIT for trivial single-tool calls where the schema is self-explanatory.

STEERING DIRECTIVES
- INCLUDE when reasoning hazards exist: tool-name confusion, parameter
  validation, mechanism-vs-outcome distinction, evidence quality.
- OMIT for trivial cases.

## SECTION PROFILE BY DOMAIN

| Section            | SWE  | Tool Use   | Customer Service | Deep Research |
|----------|---|------|---------|--------|
| ENVIRONMENT STATE  | yes  | yes (mini) | yes (data dump)  | no            |
| TASK CONTEXT       | yes  | yes        | yes              | yes           |
| TOOL SCHEMAS       | yes  | yes        | yes              | yes           |
| DOMAIN RULES       | yes  | usually no | yes              | yes           |
| STEERING DIRECTIVES| yes  | usually no | yes              | yes           |
| PREDICTION TARGET  | yes  | yes        | yes              | yes           |

## DOMAIN-SPECIFIC GUIDANCE

### SWE
- ENVIRONMENT STATE: repository root path, visible file paths from prior
  turns, note that the rest of the tree is discoverable.
- TASK CONTEXT: high-level repo/PR/feature objective; tests-vs-source
  modification policy.
- TOOL SCHEMAS: editor/shell commands with sub-commands and parameter
  validity (e.g., `view_range` only valid on files, not directories).
- DOMAIN RULES: language, test runner, "do not modify tests", project
  conventions, what the relevant module does.
- STEERING DIRECTIVES: file-slice semantics; navigation patterns; tool
  errors are determined by command/path validity, not task intent.

### Tool Use
- ENVIRONMENT STATE: keep to a small block: input record + any parsed/
  derived state. No global DB dumps.
- TASK CONTEXT: one or two lines describing the call's intent.
- TOOL SCHEMAS: full schema for the called tool: parameters, types,
  format constraints, return shape, error conditions.
- DOMAIN RULES / STEERING DIRECTIVES: usually omitted.

### Customer Service
- ENVIRONMENT STATE: data-dump style. Customer profile (id, name, DOB,
  email, phone, address, account_status, created_at, last_extension_date,
  goodwill credit), payment methods, associated lines/bills/IDs,
  retrievable fields, current timestamp. Include adjacent records that
  the next call won't read (distractors).
- TASK CONTEXT: supported task categories, identification status,
  current step in the workflow.
- TOOL SCHEMAS: lookup tools, action tools, transfer protocol with the
  exact handoff message string.
- DOMAIN RULES: policy excerpts, status semantics for accounts/lines/
  bills, scope boundaries, transfer protocol order.
- STEERING DIRECTIVES: surface behavior from tool outputs and policy,
  not assumed resolutions; status-driven branching; when to transfer.

### Deep Research
- ENVIRONMENT STATE: OMIT.
- TASK CONTEXT: research question verbatim, domain/topic area, expected
  answer type, known constraints (thresholds, exclusions, reference
  points).
- TOOL SCHEMAS: web_search / read_webpage (or equivalents) with full
  response and error shapes (e.g., `{url, error}` for failures).
- DOMAIN RULES: synthesis methodology; what evidence must establish for
  the question; how partial evidence combines.
- STEERING DIRECTIVES: source preferences (authoritative > aggregator),
  treat constraints as route/property checks rather than assumptions,
  incremental retrieval semantics.

## FORMAT AND STYLE
- Default to bullet/structured lists with sub-bullets; avoid prose
  paragraphs unless the content is genuinely narrative.
- Use data-dump format (key: value, indented sub-rows) when
  ENVIRONMENT STATE is rich (customer profiles, DB rows, repo state).
- Use exact tool, parameter, and field names from the trajectory.
- Third person, present tense, declarative voice.
- Vary phrasing across instances; do not produce templated boilerplate.
- Never quote the literal outcome; describe its shape and what determines
  it.
- The PREDICTION TARGET line is always the final line of the output, plain
  text, no `##` header, no trailing content.

## DELIBERATE NOISE INJECTION
For ENVIRONMENT STATE (when present):
- Include neighboring/irrelevant rows, columns, fields, sibling files, or
  alternative records.
- For multi-app contexts, list ALL account credentials, not only the
  active app's.
- Include extra DB rows beyond the strictly necessary ones.

For TOOL SCHEMAS:
- Include compact one-line schemas of sibling tools that could be confused
  with the active one.

This noise is intentional and trains selective attention.

## VALIDATION-FAILURE SEMANTICS
When the action omits a required parameter or violates a typed constraint
(missing required IDs, wrong sort_by prefix, non-integer values, invalid
enum, malformed dates), DOMAIN RULES must specify:
  - which parameters are required and their types/format constraints,
  - the validation-error response shape (e.g., HTTP 422 with message
    "Validation error. Reason: \n<param>: <reason>").
The predicted response is the validation error, not a successful payload.

## COMMON PITFALLS TO AVOID
- Using all 5 sections when the domain calls for fewer (Tool Use,
  Deep Research).
- Including ENVIRONMENT STATE for Deep Research.
- Skipping or moving the PREDICTION TARGET line.
- Inventing tool fields not present in the schema.
- Predicting the outcome value instead of what determines it.
- Producing identical phrasing across different trajectories.
- Forgetting noise injection in rich ENVIRONMENT STATE blocks.
- Using prose paragraphs where bullets/data-dumps are conventional.

## WORKED EXAMPLES (one per domain, abbreviated)

### Example A - Tool Use (calculate_age)

## ENVIRONMENT STATE
- Tool backing data / query state:
  - Input record:
    - date_of_birth: 1990-05-15
  - Computed / derived state:
    - date_of_birth parsed as 1990-05-15

## TASK CONTEXT
- The agent is answering the user's question about their age by calling
  the `calculate_age` tool with the provided date of birth.

## TOOL SCHEMAS
- `calculate_age`
  - Description: Calculate the age based on date of birth.
  - Parameters:
    - `date_of_birth` (string, format: date)
  - Returns:
    - `{ "age": integer }`
  - Error conditions:
    - Invalid or improperly formatted `date_of_birth` may return an error.

PREDICTION TARGET: The tool response for the `calculate_age` call in the
user turn.

### Example B - Deep Research

## TASK CONTEXT
Research Question: <verbatim user question>
Domain/Topic Area: <e.g., New Zealand urban geography>
Answer Type Expected: <e.g., a list of city names>
Known Constraints:
- <constraint 1>
- <constraint 2>

## TOOL SCHEMAS
- web_search(query: string, max_results: integer) -> {
    query, provider, results: [ {title, url, snippet} ]
  }
- read_webpage(url: string, max_chars: integer) -> {
    url, title?, content?
  }
- Failure shape: { url, error }

## DOMAIN RULES
- The agent must synthesize a defensible answer from retrieved web
  evidence.
- Relevant evidence must establish each constraint from retrieved
  sources rather than assumption.

## STEERING DIRECTIVES
- Prefer authoritative/primary sources over aggregators.
- Treat exclusion criteria as route/property checks established from
  evidence.
- Search and browsing tools return only the requested query/URL content.

PREDICTION TARGET: The tool response returned by read_webpage for the
URL and parameters in the USER turn.

### Example C - SWE

## ENVIRONMENT STATE
- Repository: `/testbed` (Python codebase).
- Visible paths: `/testbed/<module>/<file>.py`.
- Repository structure beyond the shown path is discoverable via
  filesystem tools.

## TASK CONTEXT
- Active objective: inspect/modify the repository to address the task.
- Success is evaluated against existing tests; do not modify tests.

## TOOL SCHEMAS
- `str_replace_editor`
  - `view`
    - Parameters: path: string, view_range: optional [int, int]
    - Behavior: file -> numbered slice; directory -> listing.
    - Errors: invalid path, view_range on directory, invalid line range.
  - (sibling sub-commands: `create`, `str_replace`, `insert`)

## DOMAIN RULES
- Language: Python; test runner: pytest.
- Tests under `/testbed/tests/`; do not modify them.

## STEERING DIRECTIVES
- File slice semantics: only the requested range is returned with line
  numbers.
- Tool errors depend on command/path validity, not task intent.

PREDICTION TARGET: The observation returned by `str_replace_editor` for
the requested view command.

### Example D - Customer Service

## ENVIRONMENT STATE
Current time: <timestamp>.

Customer profile:
- customer_id, full_name, date_of_birth, email, phone_number, address,
  account_status, created_at, last_extension_date,
  goodwill_credit_used_this_year

Payment methods: <list>
Associated lines: <ids>
Associated bills: <ids>

Retrievable fields via get_customer_info(): <fields>
Other domain records (lines/bills/plans/devices) accessible through
respective tools when present in the episode.

## TASK CONTEXT
- Supported task categories: technical support, overdue bill payment,
  line suspension, plan options.
- Customer identification status: <status>.

## TOOL SCHEMAS
- get_customer_by_phone / by_id / by_name
- get_details_by_id, get_bills_for_customer, get_data_usage
- suspend_line, resume_line, refuel_data, enable_roaming, disable_roaming
- send_payment_request, transfer_to_human_agents
- Transfer protocol: call transfer_to_human_agents, then send exactly
  "YOU ARE BEING TRANSFERRED TO A HUMAN AGENT. PLEASE HOLD ON."

## DOMAIN RULES
- Policy scope, status semantics (Active/Suspended/Pending/Closed for
  account/line/bill), one-bill-awaiting-payment rule, suspension lift
  rules, refuel cap (2GB), etc.

## STEERING DIRECTIVES
- Surface behavior from tool outputs and policy, not assumed resolutions.
- Status-driven branching governs eligibility for subsequent actions.
- Follow transfer protocol only when within-scope resolution is impossible.

PREDICTION TARGET: The tool response or environment response produced by
the action in the user turn.

## NOW PROCESS

[state]
<<<state>>>

[action]
<<<action>>>

[outcome (optional)]
<<<outcome>>>

Procedure:
  Step 1: Classify the domain (SWE / Tool Use / Customer Service /
          Deep Research).
  Step 2: Decide which of the conditional sections to include based on
          the inclusion rules and the per-domain profile.
  Step 3: Generate the hindsight instruction in canonical section order,
          using bullets/data-dumps as appropriate to the domain.
  Step 4: Ensure the final line is `PREDICTION TARGET: <...>`.

Output ONLY the hindsight instruction. No preamble. No postscript.
\end{verbatim}
\end{tcolorbox}

\begin{figure}
    \centering
    \includegraphics[width=0.85\linewidth]{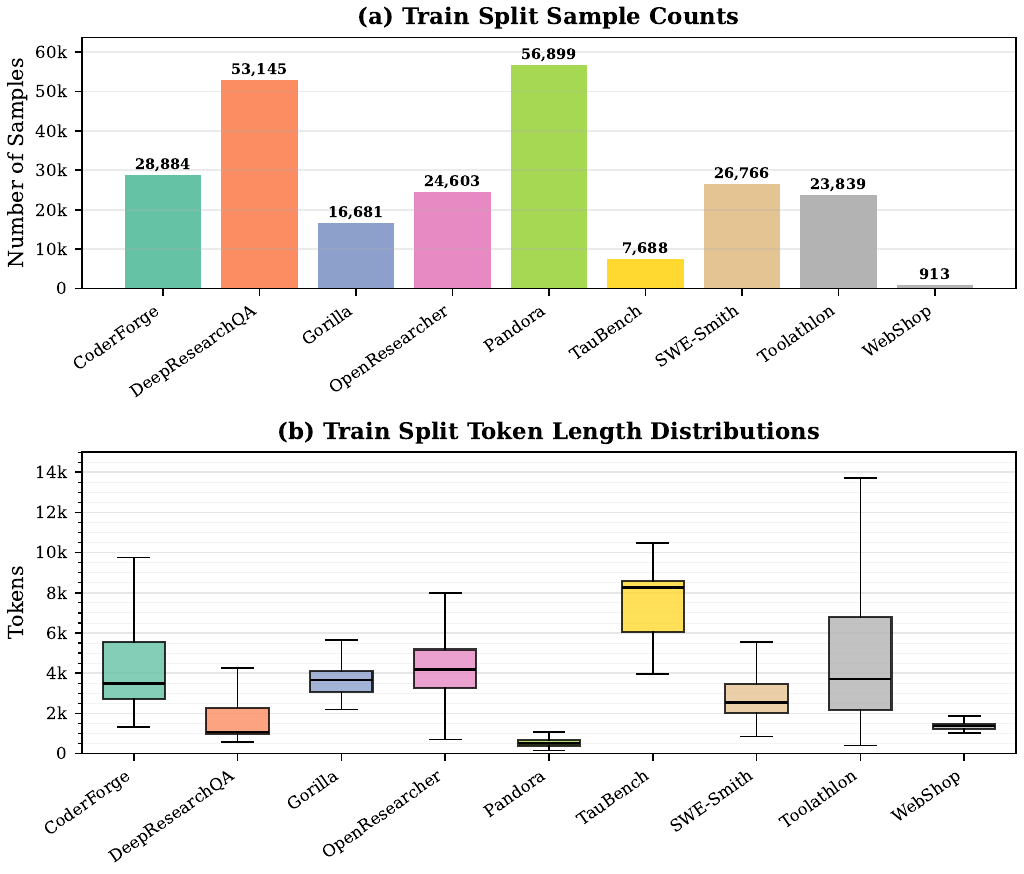}
    \caption{Training split counts and token length distributions per dataset. The sampled number of data points are proportional to openly available training tasks.}
    \label{fig:data_split_distribution}
\end{figure}

\section{Environment Setup for Downstream RL Agent Training with World Model Backend}
\label{app:rl-dataset-curation}

We describe the dataset construction and training infrastructure for each of the three interactive environments used in our experiments. In all cases, training uses a World Model (WM) for rollout simulation during GRPO, while evaluation is performed against the real environment. Reward is computed from ground-truth environment state.

\subsection{ScienceWorld}

\paragraph{Environment and Data Splits.} ScienceWorld~\cite{wang2022scienceworld} is a text-based science benchmark with 30 task types across 10 interconnected rooms, spanning state-change experiments, biological observations, and physics measurements. The agent navigates via free-text commands (\texttt{teleport to}, \texttt{pick up}, \texttt{focus on}) and submits answers via \texttt{focus on <object>}. We use 2 variations per task type for training and 2 for evaluation (60 each), drawn from disjoint variation pools. A per-task \emph{room map} produced by an environment walk visiting all 10 rooms and opening containers is baked into the WM system prompt as ground-truth object locations.

\paragraph{SFT Data and Formatting.} Expert demonstrations are generated by GPT-5.5 (\texttt{reasoning\_effort=none}) against the real environment, capped at 50 turns. Filtering to perfect-score (100) trajectories yields 37 demonstrations across diverse tasks. The agent prompt includes task-specific instructions (substance/container distinction, lifespan rankings, exploration strategy). For Mistral-7B, the system prompt is merged into the first user message, since Mistral v0.3's chat template places system messages before the \emph{last} user turn; LFM-2.5 (ChatML) and Qwen3 use their native templates.

\paragraph{Reward Structure} The reward replays agent actions against a fresh ScienceWorld instance per generation, normalized to $[{-0.2}, 1.0]$: wrong-focus penalties ($\leq -50$) clamp to $-0.2$, no-answer episodes receive $-0.15$, and mild negatives clamp to $-0.1$.

\subsection{ALFWorld}

\paragraph{Environment and Data Splits.} ALFWorld~\cite{ALFWorld20} is a TextWorld-based household environment derived from ALFRED~\cite{ALFRED20}, where the agent operates in a single room populated with receptacles (cabinets, drawers, countertops) and completes picking, placing, cleaning, heating, cooling, or examining tasks. Actions follow a fixed grammar (\texttt{go to}, \texttt{take}, \texttt{move}, \texttt{open}, \texttt{clean}, \texttt{heat}, \texttt{cool}, \texttt{use}). We use the standard splits (3{,}553 training, 140 in-distribution evaluation), sampling 300 games for RL (shuffled, seed=7) and up to 200 per task type (balanced across 6 types) for SFT.

\paragraph{SFT Data.} Expert demonstrations are generated by replaying ALFWorld's built-in handcoded expert plans against the real environment, parallelized across 8 workers and filtered to winning trajectories. This yields ${\sim}1{,}000$ demonstrations averaging 17 actions each.

\paragraph{Reward Structure} Reward is $1.0$ for goal completion (checked via trajectory predicate matching without environment access), with partial credit for pipeline steps: $0.25$ for take\,+\,place, $0.35$ for take\,+\,transform\,+\,place, and $0.03$ for take only. Episodes with 10+ actions but no take or move receive $-0.05$, and loop/repetition penalties scale reward by up to $0.3\times$. This action-oriented shaping was necessary for small models like LFM-2.5 and Qwen3 to break their instruction tuning tendencies of only interacting and exploring the environment space. These models completed approximately 50 steps before consistently giving up on exploitation heavy tasks. 

\subsection{AppWorld}

\paragraph{Environment and Data Splits.} AppWorld~\cite{appworld-acl24} is an interactive environment where agents complete tasks by making API calls to simulated applications (venmo, spotify, email, calendar, notes), receiving a natural language task description and invoking the correct sequence of tool calls. We use AppWorld's standard train/test splits, with per-task WM system prompts encoding the available API schemas and task context. The agent system prompt is shared across tasks and loaded dynamically to align SFT and RL prompts.

\paragraph{SFT Data.} Expert demonstrations are generated by running GPT-5.4 as the agent against all real AppWorld training tasks, with tool calls emitted in JSON and structured responses returned by the environment. We retain only trajectories that successfully complete the task (verified via \texttt{complete\_task}).

\paragraph{World Model and Reward.} The world model predicts API response text during GRPO rollouts, with a local proxy translating between the training plugin's payload format and the world model server's OpenAI-compatible chat completion API. The world model prompt includes the full API schema, current conversation state, and active tool call, following the same sectioned format as the other environments. Reward is computed by checking task completion predicates against the trajectory and the evaluation is done with ground truth Appworld server responses.

\subsection{Common Infrastructure}
\label{app:hyperparams}

\paragraph{Training Framework.} All models are trained using ms-swift~\cite{zhao2024swiftascalablelightweightinfrastructure} with GRPO and vLLM colocate mode for on-policy rollout generation. DeepSpeed ZeRO-2 was used for Mistral-7B-v0.3's multi-GPU training. We utilize 8xH100s for all training experiments (RL and world model) in this paper.

\paragraph{Hyperparameters.} Across all environments, we perform sweeps on learning rate, KL penalty, temperatures and rollouts. We found that the generally optimal configuration for SFT required $2 \times 10^{-6}$ to $8 \times 10^{-6}$ with LFM-2.5 and Qwen-4B models requiring a slightly higher learning rate, closer to $6 \times 10^{-6}$. For RL, we use a lower learning rate in the range of $1 \times 10^{-7}$ to $5 \times 10^{-7}$ to ensure generalization after training. We consistently use gradient accumulation of 8 steps. Furthermore, we sweep across $\{0.5, 0.7, 0.9, 1.0\}$ temperature scales following~\cite{wu2025takes, yan2025learning, Polaris2025}. We use cosine schedule for 1 epoch or convergence, whichever comes first based on the held out evaluation set. Checkpoints are saved every 20 steps and evaluated.

\section{Inference Harness and Optimizations for MDLMs}
\label{harness_optimizations}
We use \texttt{lmdeploy}~\cite{2023lmdeploy} to serve MDLMs tested in this paper. Specifically, we use a repetition penalty of 1.3 and diffusion steps \{1, 25, 50, 100\}. Since the world modeling task is deterministic and produces shorter outputs, we did not observe any output quality difference between steps 25 and 100 with the SDAR models. To maintain a balance between quality and speed, we select 50 denoising steps for our evaluations presented in~\autoref{tab:main_results}. 

While increasing block size for inference helps improve latency, we noticed a considerable performance drop when switching from a blocksize of $4 \rightarrow 8$ for the SDAR models. Since LLaDA-2.1-mini models were trained with a block size of 32, increasing their inference block size up to 64 had minimal impact on performance and accuracy in our tests when using the author recommended Joint-Threshold decoding ($\leq 5\%$ performance degradation on the test split while leading to speed-ups up to $2\times$). To ensure best performance for our results, we use the recommended block sizes per model (SDAR: 4, LLaDA-2.1-mini: 32). For WeDLM, which recommends window-based denoising instead of block-based denoising, we utilize the default window size of $W=6$ and a distance-based penalty coefficient $\lambda = 0.10$ as recommended by the authors. While this varies across different GPU architectures, we were successfully able to reproduce the inference speed figures of WeDLM of $2.5\times$ as compared to Qwen3-8B for structured and constrained tool outputs with input lengths up to 8,192 tokens on average on a single H100.

\section{Impact of thinking on AR model performance}

\begin{table}[t] \centering \small \caption{Out-of-domain evaluation of Qwen3.5-35B-A3B world model with and without thinking on \textsc{OccuBench}, \textsc{API-Bank}, and \textsc{InterCode} test splits. Reported latency is for 4xH200 GPUs with LLM baselines using vLLM + FlashInfer, SDAR using the author recommended JetEngine and WeDLM using the official, vLLM-modified inference code.} \label{tab:thinking-ablation} \begin{tabular}{lcccc} \toprule \textbf{Configuration} & \textbf{BLEU-1} & \textbf{ROUGE-L} & \textbf{MAUVE} & \textbf{Latency ($N=4567$)}\\ \midrule Qwen3.5-35B-A3B (no thinking) & .621 & .664 & .899 & 211s ($1.00\times$)\\ + Thinking ($\textsc{Effort}=5\text{k}$) & .713 & \textbf{.768} & \textbf{.940} & 793s ($3.75\times$)\\ + Thinking ($\textsc{Effort}=10\text{k}$) & .710 & .758 & .934 & 1028s ($4.87\times$)\\ 
\midrule

SDAR-8B (no thinking) & \textbf{.749} & .740 &  .902 & 380s ($1.80\times$) \\
WeDLM-8B (no thinking) & .729 & \textbf{.761} & .910 & 145s ($0.68\times$) \\

\bottomrule \end{tabular} \end{table}

As discussed in~\autoref{sec:mdlm_for_wm}, enabling thinking mode for these models has previously been shown~\cite{snell2024scaling} to improve diversity. We present a small ablation in~\autoref{tab:thinking-ablation} to show that test time thinking helps improve AR performance which can be attributed to the diversity produced by the chain of thought. This further strengthens the argument for using MDLMs for world modeling because despite being much smaller, they offer much better diversity and hence performance to latency ratios than LLMs with similar performance.   

\section{Prompts used for World Model for Downstream RL experiments}
\label{appsec:downstream_wm_prompts}

Template variables are shown in \texttt{\{curly\_braces\}}. These are filled at runtime from the per-task WM system prompt, replayed state, and current action.

\begin{tcolorbox}[title=ALFWorld World Model Prompt, colback=gray!5, colframe=gray!50, breakable]

\noindent\textbf{ENVIRONMENT STATE}

\smallskip

ALFWorld is a TextWorld household environment derived from ALFRED. The agent operates in a single room (kitchen / bathroom / bedroom / living-room / office) populated with receptacles (cabinet, drawer, fridge, microwave, countertop, shelf, sidetable, diningtable, desk, dresser, bed, sinkbasin, stoveburner, toilet, bathtub, garbagecan, etc.) each carrying a 1-indexed instance id (e.g.\ \texttt{'cabinet 8'}). Objects are also 1-indexed within their class (e.g.\ \texttt{'apple 1'}, \texttt{'mug 2'}). All names are lowercase.

\smallskip

\noindent Task: \texttt{\{task\}}\\
Task type: \texttt{\{task\_type\}}\\
Goal predicates: \texttt{\{goal\_pred\}}\\
Current location: \texttt{\{current\_loc\}}\\
Inventory: \texttt{\{inventory\}}

\smallskip

\noindent Receptacles in this room (all 1-indexed, lowercase):
\begin{itemize}
  \item \texttt{\{recep\_1\}}: \texttt{\{open\_closed\_status\}}
  \item \texttt{\{recep\_2\}}: \texttt{\{open\_closed\_status\}}
  \item \ldots
\end{itemize}

\noindent Known receptacle contents (pre-baked from initial scene walk + updates from this rollout's take/put events):

\smallskip

\hspace*{1em}\texttt{[\{recep\_1\}] \{obj\_1\}, \{obj\_2\}}\\
\hspace*{1em}\texttt{[\{recep\_2\}] (empty)}\\
\hspace*{1em}\ldots

\smallskip

\noindent Object state modifiers acquired earlier in this rollout:
\begin{itemize}
  \item \texttt{\{obj\}}: \{clean, hot, cool, sliced, on\}
\end{itemize}

\medskip

\noindent\textbf{TASK CONTEXT}

\smallskip

Predict ONLY the next textworld feedback string for the action below. The agent is interacting with the env one action at a time; you are predicting what textworld would print as the immediate response to the action -- exactly one short paragraph of plain English. Do NOT plan, do NOT solve the task, do NOT echo the action, do NOT list admissible commands, do NOT output JSON or markdown.

\smallskip

\noindent Active action (verbatim, lowercase): \texttt{\{action\_text\}}

\medskip

\noindent\textbf{DOMAIN RULES}

\smallskip

Response must match the template VERBATIM. Always include instance numbers (e.g.\ \texttt{'stoveburner 1'} not \texttt{'stoveburner'}, \texttt{'pot 1'} not \texttt{'pot'}).

\smallskip

\noindent Example: \texttt{go to stoveburner 1} $\rightarrow$ ``You arrive at stoveburner 1. On the stoveburner 1, you see a pot 1, and a pan 2.''

\medskip

\noindent\texttt{go to <recep>}
\begin{itemize}
  \item CLOSED: ``You arrive at <recep>. The <recep> is closed.''
  \item OPEN+items: ``You arrive at <recep>. On the <recep>, you see a X 1, a Y 2, and a Z 1.''
  \item OPEN+empty: ``You arrive at <recep>. On the <recep>, you see nothing.''
  \item Unknown: ``Nothing happens.''
\end{itemize}

\noindent\texttt{open <recep>}
\begin{itemize}
  \item Only valid when \texttt{<recep>} is currently CLOSED \emph{and} agent is at \texttt{<recep>}.
  \item If valid AND \texttt{<recep>} contains items: ``You open the <recep>. The <recep> is open. In it, you see <listed\_items>.''
  \item If valid AND \texttt{<recep>} is empty: ``You open the <recep>. The <recep> is open. In it, you see nothing.''
  \item Otherwise (already open, not openable, or not at it): ``Nothing happens.''
\end{itemize}

\noindent\texttt{close <recep>}
\begin{itemize}
  \item Only valid when \texttt{<recep>} is currently OPEN (and was openable in the first place) and agent is at \texttt{<recep>}.
  \item Success: ``You close the <recep>.''
  \item Otherwise: ``Nothing happens.''
\end{itemize}

\noindent\texttt{take <obj> from <src>}
\begin{itemize}
  \item Valid iff agent is at \texttt{<src>}, \texttt{<src>} is OPEN, and \texttt{<obj>} is currently in \texttt{<src>}.
  \item Success: ``You pick up the <obj> from the <src>.''
  \item \texttt{<obj>} MUST include the instance number: ``You pick up the pot 1'' not ``You pick up the pot''.
  \item Otherwise: ``Nothing happens.''
  \item After this, \texttt{<obj>} moves from \texttt{<src>} to inventory.
\end{itemize}

\noindent\texttt{put <obj> in/on <dst>} and \texttt{move <obj> to <dst>} (interchangeable)
\begin{itemize}
  \item Valid iff \texttt{<obj>} is in inventory, agent is at \texttt{<dst>}, \texttt{<dst>} is OPEN.
  \item Success: ``You move the <obj> to the <dst>.'' (always `move \ldots\ to', never `put \ldots\ in/on', regardless of which form the agent typed.)
  \item Otherwise: ``Nothing happens.''
\end{itemize}

\noindent\texttt{clean <obj> with <recep>} (recep must be a sinkbasin)
\begin{itemize}
  \item Valid iff \texttt{<obj>} is in inventory, agent is at \texttt{<recep>}, \texttt{<recep>} is a sinkbasin.
  \item Success: ``You clean the <obj> using the <recep>.'' Object gains `clean'.
  \item Otherwise: ``Nothing happens.''
\end{itemize}

\noindent\texttt{heat <obj> with <recep>} (recep must be a microwave)
\begin{itemize}
  \item Valid iff \texttt{<obj>} is in inventory, agent is at \texttt{<recep>}, \texttt{<recep>} is a microwave.
  \item Microwaves do NOT need to be opened to heat -- heat is a permitted action even when the microwave is closed.
  \item Success: ``You heat the <obj> using the <recep>.'' Object gains `hot'.
  \item If the agent then heats again or cools, the latest action wins (sticky).
  \item Otherwise: ``Nothing happens.''
\end{itemize}

\noindent\texttt{cool <obj> with <recep>} (recep must be a fridge)
\begin{itemize}
  \item Valid iff \texttt{<obj>} is in inventory, agent is at \texttt{<recep>}, \texttt{<recep>} is a fridge.
  \item Fridges do NOT need to be opened to cool.
  \item Success: ``You cool the <obj> using the <recep>.'' Object gains `cool'.
  \item Otherwise: ``Nothing happens.''
\end{itemize}

\noindent\texttt{slice <obj> with <tool>} (tool must be a knife)
\begin{itemize}
  \item Valid iff \texttt{<tool>} is in inventory and \texttt{<obj>} is at the agent's location or in inventory.
  \item Success: ``You slice the <obj> using the <tool>.'' Object gains `sliced'.
  \item Otherwise: ``Nothing happens.''
\end{itemize}

\noindent\texttt{use <obj>} (typically \texttt{use desklamp N})
\begin{itemize}
  \item Valid iff \texttt{<obj>} is at the agent's location or in inventory. \texttt{<obj>} is an object (not a receptacle) -- desklamps, floorlamps, etc.\ are not listed in the receptacle table.
  \item Success: ``You turn on the <obj>.'' Object gains `on'.
  \item Otherwise: ``Nothing happens.''
\end{itemize}

\noindent\texttt{examine <thing>} -- read-only; never changes the world.
\begin{itemize}
  \item If \texttt{<thing>} is a receptacle at the agent's location: ``On the <thing>, you see <listed\_items>.'' (or `\ldots you see nothing.')
  \item If \texttt{<thing>} is an object in inventory or visible at current location: ``This is a <state\_words> <thing>.'' where \texttt{<state\_words>} reflects any modifiers (cold, hot, clean, sliced) -- e.g.\ ``This is a cold tomato 1.'' If no modifier applies: ``There's nothing special about <thing>.''
  \item Otherwise: ``Nothing happens.''
\end{itemize}

\noindent\texttt{look}
\begin{itemize}
  \item If agent is in the middle of the room (not at any receptacle): ``You are in the middle of a room. Looking quickly around you, you see nothing.'' (Note: textworld emits the receptacle list ONLY in the reset welcome message, never on a subsequent \texttt{look}.)
  \item If agent is facing a receptacle: ``You are facing the <recep>. Next to it, you see <adjacent\_items\_or\_nothing>.''
  \item Adjacent receptacles in the same fixture group may also be listed: ``You are facing the shelf 2, and shelf 1. Next to it, you see \ldots''.
\end{itemize}

\noindent\texttt{inventory}
\begin{itemize}
  \item If carrying nothing: ``You are not carrying anything.''
  \item Otherwise: ``You are carrying: <listed\_items>.'' using the same Oxford-comma format as receptacle contents.
\end{itemize}

\noindent\texttt{help} -- emits the verbatim TextWorld command-list block (rare; agent should not use this).

\medskip

\noindent\textbf{Universal fallback:} any action whose preconditions are not satisfied -- wrong location, missing object, closed destination for put/move, unknown object id, unknown receptacle id, malformed verb -- produces exactly:

\smallskip

\hspace*{1em}``Nothing happens.''

\smallskip

\noindent (Two words, capital N, period at end. No additional context.)

\medskip

\noindent\textbf{Closedness rules} (which classes default to closed):
\begin{itemize}
  \item Default-closed: cabinet, drawer, fridge, microwave, safe, box. Must be opened before take/put/move using their interior. Heat/cool/clean do NOT require open.
  \item Default-open / not-openable: countertop, shelf, sidetable, diningtable, desk, dresser, bed, sinkbasin, stoveburner, toaster, coffeemachine, toilet, bathtub, sofa, armchair, garbagecan, ottoman, tvstand, hand-/towel-holder, toiletpaperhanger.
  \item Use the receptacle table above as the authoritative open/closed source for this specific game; it overrides the class defaults when stated.
\end{itemize}

\medskip

\noindent\textbf{STEERING DIRECTIVES}
\begin{itemize}
  \item Copy receptacle and object names EXACTLY from Environment State (e.g.\ ``desk 1'' not ``table'', ``sinkbasin 1'' not ``sink'').
  \item Every name must include its instance number.
  \item Only use objects/receptacles from the Environment State. Do not invent.
  \item If uncertain, output ``Nothing happens.''
  \item One short paragraph. No markdown, JSON, or admissible-commands lists.
  \item Prefix each item with ``a '' in listings (e.g.\ ``a apple 1, a mug 2, and a pan 1'').
\end{itemize}

\medskip

\noindent\textbf{PREDICTION TARGET:} The single textworld feedback paragraph that the engine would print in response to the action \texttt{\{action\_text\}} from the environment state above.

\end{tcolorbox}

%%%%%%%%%%%%%%%

\begin{tcolorbox}[title=Appworld World Model Prompt, colback=gray!5, colframe=gray!50, breakable]

\noindent\textbf{ENVIRONMENT STATE}

\smallskip

The environment is AppWorld tool-response prediction.

\smallskip

\noindent Active task:
\begin{itemize}
  \item \texttt{\{active\_task\}}
\end{itemize}

\noindent Current date:
\begin{itemize}
  \item \texttt{\{current\_date\}}
\end{itemize}

\noindent Account credentials:
\begin{itemize}
  \item \texttt{\{app\}}: username=\texttt{\{username\}}, password=\texttt{\{password\}}
  \item \ldots
\end{itemize}

\noindent Active tool call:
\begin{itemize}
  \item tool\_name: \texttt{\{tool\_name\}}
  \item arguments: \texttt{\{tool\_args\_json\}}
  \item validation\_status: \texttt{\{valid|invalid\}}
  \item validation\_error: \texttt{\{error\_message|null\}}
\end{itemize}

\medskip

\noindent\textbf{TASK CONTEXT}

\smallskip

Predict only the JSON tool response for this one AppWorld action. Do not solve the whole user task. Do not echo the action.

\medskip

\noindent\textbf{TOOL SCHEMAS}
\begin{itemize}
  \item \texttt{\{tool\_name\}(\{param1\}, \{param2\}, \ldots)}
  \item Return shape for list/search tools: \texttt{\{"total": int, "\{return\_key\}": array\}}
  \item Relevant source section: \texttt{\{section\}}
  \item For invalid calls, return exactly one JSON object with an \texttt{error} key.
\end{itemize}

\medskip

\noindent\textbf{DOMAIN RULES}
\begin{itemize}
  \item Validate the tool name and argument value types before generating a response.
  \item If \texttt{validation\_status} is valid, the active tool is available and the arguments passed validation.
  \item [If invalid]: Unknown tools, bad credentials, schema/type placeholder arguments, and missing target records return an error JSON object.
  \item [If valid]: This is not an unknown-tool case. Do not return an error for tool availability.
  \item Returned records and IDs must come only from the record lines shown below.
  \item Return strict valid JSON only. Use JSON \texttt{true}/\texttt{false}/\texttt{null}, not Python \texttt{True}/\texttt{False}/\texttt{None}.
  \item No markdown, no commentary, no tracebacks, and do not double-quote or escape the whole JSON object.
  \item Do not include \texttt{<think>} text. Wrap the final JSON in \texttt{<tool\_response>} only if the model requires tags.
\end{itemize}

\medskip

\noindent\textbf{COMPUTED QUERY STATE}
\begin{itemize}
  \item matched\_count\_after\_filters: \texttt{\{matched\_count\}}
  \item page\_index: \texttt{\{page\_index\}}
  \item page\_limit: \texttt{\{page\_limit\}}
  \item page\_offset: \texttt{\{page\_offset\}}
  \item records\_on\_this\_page: \texttt{\{num\_records\}}
  \item The response \texttt{total} field must equal \texttt{matched\_count\_after\_filters}.
  \item The response array must contain only the \texttt{records\_on\_this\_page} listed below.
\end{itemize}

\noindent Strict JSON records for this response page:

\smallskip

\hspace*{1em}\texttt{\{json\_records\}}

\medskip

\noindent\textbf{STEERING DIRECTIVES}
\begin{itemize}
  \item Expected validation status: \texttt{\{valid|invalid\}}.
  \item [If invalid]: The exact error response body is: \texttt{\{"error": "\{error\_message\}"\}}.
  \item [If valid]: The tool call is valid; do not return an error.
  \item For list/search tools, apply filters before pagination; never repeat page 0 for later pages.
  \item For login success, return a short access token only when credentials match Account credentials.
  \item If this is a list/search success, return \texttt{\{"total": \{total\}, "\{return\_key\}": [\ldots]\}} using only Strict JSON records for this response page.
\end{itemize}

\medskip

\noindent\textbf{PREDICTION TARGET:} The strict JSON tool response for the action in the user turn.

\medskip
\hrule
\medskip

\noindent\textbf{Notes}

\smallskip

The AppWorld WM prompt is substantially more structured than the other two environments because:

\begin{enumerate}
  \item \textbf{Tool call validation} is performed locally before the WM is called. Invalid tool names, bad credentials, and schema-as-argument errors are caught and returned as deterministic error JSON without WM involvement.
  \item \textbf{Query pre-computation}: For list/search tools (\texttt{spotify\_\_search\_songs}, \texttt{venmo\_\_show\_transactions}, etc.), the local responder filters records from the baked system prompt, applies pagination, and provides the exact JSON records the WM should return. This reduces the WM's job to formatting -- it just needs to wrap the pre-computed records in the correct response shape.
  \item \textbf{Per-tool routing}: The \texttt{SECTION\_BY\_TOOL} mapping determines which section of the system prompt to query (songs, albums, transactions, etc.), and \texttt{RETURN\_KEY\_BY\_SECTION} determines the JSON key for the response array.
  \item \textbf{Mutation tools} (create, delete, like, unlike, etc.) are handled deterministically with \texttt{\{"status": "success", "message": "Action completed"\}} -- the WM is not called for these.
\end{enumerate}

\end{tcolorbox}

\begin{tcolorbox}[title=ScienceWorld World Model Prompt, colback=gray!5, colframe=gray!50, breakable]

\noindent\textbf{ENVIRONMENT STATE}

\smallskip

ScienceWorld is a text-based interactive science experiment environment with 10 rooms.

\smallskip

\noindent The agent is currently in: \texttt{\{current\_room\}}.\\
Task: \texttt{\{task\_desc\}}

\smallskip

\noindent Visible objects in \texttt{\{current\_room\}}:

\smallskip

\hspace*{1em}\texttt{\{room\_contents\}}

\smallskip

\noindent Agent's inventory: \texttt{\{inventory\_items\}}

\smallskip

\noindent Room adjacency (for `go to' validation):

\smallskip

\hspace*{1em}\texttt{\{room\_connections\}}

\medskip

\noindent\textbf{TASK CONTEXT}

\smallskip

The agent navigates rooms, picks up objects, and performs science experiments. Actions are plain text commands like `teleport to kitchen', `pick up glass cup', `focus on water'.

\smallskip

\noindent The active action is: \texttt{\{action\_text\}}

\medskip

\noindent\textbf{DOMAIN RULES}
\begin{itemize}
  \item \texttt{teleport to X} always succeeds. Response: ``You teleport to the X.''
  \item \texttt{go to X} only works if X is adjacent to the current room. If not adjacent: ``No known action matches that input.''
  \item \texttt{look around} returns a plain text list of objects in the current room. Format: ``This room is called X. In it, you see: obj1, obj2, obj3. You also see: door to Y, door to Z.''
  \item \texttt{pick up X} works only if X is visible in the current room AND has not already been picked up. Response: ``You move the X to the inventory.'' If X is not present or already picked up: ``No known action matches that input.''
  \item \texttt{focus on X} works ONLY if X is explicitly listed in the current room description OR was previously picked up (in inventory). If X does NOT appear in the visible objects or inventory, respond: ``No known action matches that input.'' NEVER confirm focus on an object that is not visibly present or in inventory -- this is critical.
  \item \texttt{activate X} / \texttt{deactivate X} works if X is in the current room. Response: ``The X is now activated/deactivated.''
  \item \texttt{put X in/on Y} works if X is in inventory. Response: ``You move the X to the Y.''
  \item \texttt{open X} / \texttt{close X} works on doors and containers in the current room.
  \item \texttt{inventory} lists items the agent has picked up.
  \item \texttt{wait} / \texttt{wait1} advances time. Response: ``(1 tick passes)''
  \item Invalid actions or objects not present: ``No known action matches that input.''
  \item Items picked up earlier in the conversation are in inventory and no longer in their original room.
\end{itemize}

\medskip

\noindent\textbf{STEERING DIRECTIVES}

\smallskip

\noindent \textbf{GROUNDING RULE:} Your response MUST use ONLY objects, rooms, and substances that appear in the ENVIRONMENT STATE section above. Do NOT invent, hallucinate, or elaborate beyond what is listed. If the environment state lists `a brick, a workbench', respond with EXACTLY those items -- do not add tools, equipment, or descriptions that are not in the state.

\begin{itemize}
  \item For \texttt{look around}: copy the room description from the Visible objects section above VERBATIM. Do not add objects, do not elaborate, do not write prose. Just the object list.
  \item For \texttt{pick up X}: check if X appears in the Visible objects. If yes: ``You move the X to the inventory.'' If no: ``No known action matches that input.''
  \item For \texttt{focus on X}: check Visible objects AND inventory. If X not found: ``No known action matches that input.''
  \item Keep responses SHORT -- one or two sentences maximum. Never write paragraphs, stories, or elaborate descriptions.
  \item Output plain text only. No JSON, no markdown, no code, no arrays, no role tags.
\end{itemize}

\medskip

\noindent\textbf{PREDICTION TARGET:} The ScienceWorld observation text produced by the action \texttt{`\{action\_text\}'} given the current environment state.

\end{tcolorbox}

\section{World Model Failure Modes}
\label{appsec:wm-failures}

\newcommand{\trajsnip}[1]{%
  \par\vspace{2pt}%
  {\scriptsize\ttfamily\raggedright #1\par}%
}

\begin{table}[h]
\centering
\small
\caption{World model failure modes observed during GRPO training.
\textcolor{red}{Red} highlights the erroneous portion.
AW = AppWorld, ALF = ALFWorld, SW = ScienceWorld. These failures were
addressed via grounding rules in the WM prompt, local deterministic
responders, and JSON unwrapping in the training plugin.}
\label{tab:wm-failures}
\setlength{\tabcolsep}{4pt}
\renewcommand{\arraystretch}{1.2}
\begin{tabularx}{\linewidth}{@{}
    >{\raggedright\arraybackslash}p{2.6cm}
    >{\centering\arraybackslash}p{0.9cm}
    >{\raggedright\arraybackslash}X@{}}
\toprule
\textbf{Failure Mode} & \textbf{Env} & \textbf{Trajectory Snippet} \\
\midrule

(i) Incorrect pagination & AW &
Agent requests page~1 of recommendations; WM returns the same records as
page~0. Downstream, the agent double-counts songs and computes incorrect
aggregates.
\trajsnip{%
T4: show\_recommendations(\{page\_index:0\}) $\to$ \{total:162, songs:[\{id:40\},\{id:41\},\dots]\}\newline
T5: show\_recommendations(\{page\_index:1\}) $\to$ \{total:162, songs:[\{id:40\},\{id:41\},\dots]\} \textcolor{red}{$\leftarrow$ same IDs}%
} \\
\addlinespace

(ii) Object hallucination & SW &
WM invents code-execution errors and objects absent from the environment
state. Agent receives fabricated feedback that does not correspond to any
valid ScienceWorld response.
\trajsnip{%
Agent: open cupboard\newline
WM: ``Error during execution: name `cupboard\_is\_opened' is not defined'' \textcolor{red}{$\leftarrow$ fabricated}\newline[2pt]
Agent: activate solar panel\newline
WM: ``Solar panels cannot be used directly.'' \textcolor{red}{$\leftarrow$ not a ScienceWorld response}%
} \\
\addlinespace

(iii) JSON-wrapped responses & SW &
WM outputs structured JSON instead of plain-text environment observations.
Downstream parser receives role/content wrappers that corrupt the
conversation history.
\trajsnip{%
Agent: look around\newline
WM: [\{``role'':``tool'', ``content'':``This room is called the kitchen. In it, you see: the agent, a substance called air, a chair\dots''\}] \textcolor{red}{$\leftarrow$ JSON wrapper}%
} \\
\addlinespace

(iv) Dropped instance numbers & ALF &
WM substitutes generic class names for specific receptacle instances,
breaking downstream state tracking and action validation.
\trajsnip{%
Agent: go to toaster 1\newline
WM: ``You arrive at toaster 1. On the \textcolor{red}{table}, you see a bread 2.''\newline
\textcolor{gray}{Expected: ``On the \textbf{toaster 1}, you see a bread 2.''}\newline[2pt]
Agent: go to microwave 1\newline
WM: ``You arrive at microwave 1. On the \textcolor{red}{table}, nothing\dots''\newline
\textcolor{gray}{Expected: ``On the \textbf{microwave 1}, you see nothing.''}%
} \\
\addlinespace

(v) State inconsistency & SW &
WM confirms picking up an object that is already in inventory, creating
phantom duplicates. Agent ``picks up'' the thermometer three times across
17 turns.
\trajsnip{%
T6: pick up thermometer $\to$ ``You move the thermometer to the inventory.''\newline
T15: pick up thermometer $\to$ ``You move the thermometer to the inventory.'' \textcolor{red}{$\leftarrow$ already held}\newline
T17: pick up thermometer $\to$ ``You move the thermometer to the inventory.'' \textcolor{red}{$\leftarrow$ third time}%
} \\
\addlinespace

(vi) Verbose / garbled output & ALF, SW &
WM generates excessively long descriptions with hallucinated items, or
garbles object names by dropping instance numbers from object listings.
\trajsnip{%
\textbf{ALFWorld:}\newline
Agent: go to fridge 1\newline
WM: ``\dots On the fridge, you see a \textcolor{red}{2}, and a mug.'' \textcolor{gray}{(object class dropped)}\newline[2pt]
\textbf{ScienceWorld (Qwen WM):}\newline
Agent: look around (greenhouse)\newline
WM: ``\dots a decorative stone gazing ball resting atop one corner\dots\ a rustic lantern holding dried lavender bundles\dots\ finally tucked away quietly within deepest recesses thereof lies \textcolor{red}{forgotten treasure trove comprising ancient maps}\dots'' \textcolor{gray}{(886 hallucinated tokens)}%
} \\

\bottomrule
\end{tabularx}
\end{table}

\autoref{tab:wm-failures} catalogs recurring world model (WM) failure modes
observed during GRPO training for which prompts were later optimized. We present trajectory snippets from actual training
runs to help explainability. These failures led to degradation of RL training signal and caused agent policy collapse for specific categories of tasks on the held out test set. We observed similar issues with both MDLMs and AR models but the hallucination and state inconsistency patterns for AR models were much more evident and harmful to agent behavior generalization.

\section{Full three-shot and non-Transformer results.}
\label{app-sec:additonal-results}
Table~\ref{tab:three_shot_results} reports three-shot results for all baseline models and for the finetuned SDAR models; Table~\ref{tab:main_results} retains only the best-performing three-shot AR (GLM-4.7-Flash) and MDLM (SDAR-30B-A3B) baselines for brevity. Across every column, the finetuned SDAR-30B-A3B remains the strongest model under three-shot prompting. To isolate the diffusion objective from architectural choices, Table~\ref{tab:non_standard_arch} additionally reports finetuned non-Transformer baselines, RWKV-7B-g1 and the Mamba-attention hybrid Falcon-H1R-7B, trained with their official harnesses using the setup in Appendix~D.4. Both underperform the size-matched AR baseline (Qwen-3-8B), which in turn underperforms SDAR-8B, indicating that the gains reported in Table~\ref{tab:main_results} do not arise from fixed-state or linear-attention alternatives to autoregressive decoding.

\begin{table}[t]
  \centering
  \caption{Full three-shot results for all baseline and finetuned models on in-domain and out-of-domain test sets. Best three-shot baseline result in each column is \textbf{bolded}; best finetuned three-shot result is bold highlighted in \textbf{\textcolor{red}{red}}. The best-performing 3-shot AR and MDLM baselines (GLM-4.7-Flash and SDAR-30B-A3B) are also reported in Table~\ref{tab:main_results}.}
  \label{tab:three_shot_results}
  \footnotesize
  \setlength{\tabcolsep}{3pt}
  \resizebox{\textwidth}{!}{%
  \begin{tabular}{l ccc ccc ccc ccc}
    \toprule
    & \multicolumn{3}{c}{\textbf{In-domain}}
    & \multicolumn{3}{c}{\textbf{API-Bank}}
    & \multicolumn{3}{c}{\textbf{OccuBench}}
    & \multicolumn{3}{c}{\textbf{Intercode (SQL)}} \\
    \cmidrule(lr){2-4} \cmidrule(lr){5-7} \cmidrule(lr){8-10} \cmidrule(lr){11-13}
    \textbf{Model}
    & B-1 & R-L & MAUVE
    & B-1 & R-L & MAUVE
    & B-1 & R-L & MAUVE
    & B-1 & R-L & MAUVE \\
    \midrule
    \multicolumn{13}{l}{\textit{3-shot Baseline Models}} \\
    \midrule
    Qwen-3-8B  & .379 & .506 & .348 & .578 & .509 & .365 & .559 & .544 & .788 & .532 & .523 & .833 \\
    Qwen-3.5-27B       & \textbf{.509} & .578 & .588 & .611 & .660  & .600 & .731 & .772 & .834 & .585 & .623 & .849 \\
    Qwen-3.5-35B-A3B   & .334 & .507 & .590 & .621 & .663 & .634 & .744 & \textbf{.781} & .832 & .589 & \textbf{.662} & .851 \\
    % Nemotron-3-Nano-30B& .346 & .555 & .618 & .675 & .667 & .700 & .738 & .724 & .845 & .584 & .551& .825 \\
    GLM-4.7-Flash      & .401 & .565 & .629 & \textbf{.686} & \textbf{.689}  & \textbf{.707} & \textbf{.746}  & .742 & .838 & .612 & .597 & .822 \\
    % GPT-OSS-20B        & .339 & .555 & .580 & .662 & .670 & .699 & .731 & .720 & .822 & .644 & .640 & .880 \\
    LLaDA-2.1-mini     & .390 & .521 & .523 & .567 & .589 & .555 & .655 & .613 & .819 & .425 & .462 & .821 \\
    WeDLM-8B & .382 & .512 & .400 & .582 & .511 & .451  & .692 & .755 & .826 & .568 & .555 & .866 \\
    SDAR-8B & .349 & .451 & .500 & .588 & .619 & .668 & .621 & .729 & \textbf{.945} & .611 & .622 & .870\\
    SDAR-30B-A3B & .401 & \textbf{.601} & \textbf{.666} & .601 & .662 & .702 & .700 & .713 & .849 & \textbf{.621} & .641 & \textbf{.886}\\
    \midrule
    \multicolumn{13}{l}{\textit{3-shot Finetuned Diffusion Models}} \\
    \midrule
    SDAR-8B & .891 & .899 & .990 & .946 & .889 & .882 &  .831 & .810 & .981 & .822 & .824 & .967 \\
    SDAR-30B-A3B & \textbf{\textcolor{red}{.905}} & \textbf{\textcolor{red}{.910}} & \textbf{\textcolor{red}{.995}} & \textbf{\textcolor{red}{.951}} & \textbf{\textcolor{red}{.954}} & \textbf{\textcolor{red}{.900}} & \textbf{\textcolor{red}{.904}} & \textbf{\textcolor{red}{.862}} & \textbf{\textcolor{red}{.983}} & \textbf{\textcolor{red}{.831}} & \textbf{\textcolor{red}{.834}} & \textbf{\textcolor{red}{.970}} \\
    \bottomrule
  \end{tabular}%
  }
\end{table}

\begin{table}[t]
  \centering
  \caption{Generation quality of finetuned non-Transformer sequence models compared to the AR and MDLM baselines closest in parameter count. Scores are averaged over the three out-of-domain test sets (API-Bank, OccuBench, Intercode-SQL); each model is finetuned with its official training harness and we report the best of three hyperparameter configurations (Appendix~D.4). The Qwen-3-8B and SDAR-8B rows are the corresponding OOD averages of the finetuned models in Table~\ref{tab:main_results}. Best result in each column is \textbf{bolded}.}
  \label{tab:non_standard_arch}
  \footnotesize
  % \setlength{\tabcolsep}{6pt}
  % \resizebox{\linewidth}{!}{%
  \begin{tabular}{l l ccc}
    \toprule
    \textbf{Model} & \textbf{Architecture} & B-1 & R-L & MAUVE \\
    \midrule
    RWKV-7B-g1     & RWKV (linear attention) & .334 & .496 & .651 \\
    Falcon-H1R-7B  & Mamba-attention hybrid & .541 & .661 & .807 \\
    Qwen-3-8B      & Transformer (AR)        & .636 & .682 & .873 \\
    SDAR-8B        & Transformer (MDLM)      & \textbf{.749} & \textbf{.740} & \textbf{.902} \\
    \bottomrule
  \end{tabular}%
  % }
\end{table}

% ==============

\section{Additional Ablations, Analyses and Adjacent Research Questions}
\label{app:additional_rqs}
 
\paragraph{Separating the diffusion objective from the attention pattern.}
SDAR and LLaDA-2.1 use block-causal attention: attention is bidirectional within a block and causal across blocks. WeDLM-8B keeps Qwen3-8B's strictly causal (lower-triangular) mask. It moves already-decoded tokens to the front of the input but keeps their original position IDs, so a masked slot can attend to tokens that come later in the sequence without any bidirectional attention. Fine-tuned WeDLM-8B beats fine-tuned Qwen3-8B on 11 of 12 split-metric pairs (~\autoref{tab:main_results}), so bidirectional attention alone cannot explain the gain. Two caveats apply in this case: first, WeDLM had additional continued pre-training on top of Qwen3-8B, so part of the gap may come from extra data. Second, its streaming decoder commits a left-to-right prefix under a distance penalty. We therefore treat WeDLM only as a control for the attention pattern and nothing else. Our evidence for anchor-aware decoding is the SDAR study below.
 
\paragraph{Why iterative denoising rather than a one-shot bidirectional fill.}
A single-pass (BERT-style) fill treats masked positions as conditionally independent given the context, so it cannot resolve dependencies among unknown fields. In~\autoref{fig:mdlm_vs_ar}, for example, the service and fee fields both depend on the depot the model has not yet generated. A one-shot fill is the maximum-parallelism end of the decoding-step axis. Prior MDLM work finds that quality falls as parallelism rises and that errors from committing tokens in parallel propagate downstream \citep{cheng2025sdar,bie2025llada2,liu2025wedlm}.Diffusion heads can be attached to encoder-only, encoder-decoder \citep{arriola2026encoder}, or decoder-only backbones. We study AR-adapted MDLMs because they are the strongest instruction-following MDLMs available, and adapting from an AR model carries over its pre- and post-training \citep{gong2025scalingdiffusionlanguagemodels}.
 
\paragraph{Synthetic anchor benchmark.}
\label{app:anchor}
To measure the anchor effect sketched in Figure~1, we take 250 OOD samples, decode deterministically, and fix $k\in\{1,2,8,16\}$ future anchor tokens in the target state. For SDAR-8B we use the block-size-32 checkpoint and place all $k$ anchors inside one block. For the AR baseline (Qwen3.5-27B) we place them anywhere in the target. We report accuracy (Table~\ref{tab:anchor}). The MDLM-AR gap widens monotonically from $+0.04$ ($k{=}1$) to $+0.08$ ($k{=}16$). The AR model stays at $0.86$ because it cannot condition on anchors placed after the position it is generating, even though it is over $3\times$ larger. Letting SDAR commit the anchors itself in the first denoising step gives similar accuracy (0.89-0.91).
 
\begin{table}[h]
\centering\small
\caption{Synthetic anchor benchmark (250 OOD samples). $k$: number of future anchors fixed in the target state.}
\label{tab:anchor}
\begin{tabular}{lccc}
\toprule
$k$ & Qwen3.5-27B (AR) & SDAR-8B (forced) & SDAR-8B (self-committed) \\
\midrule
1  & .86 & .90 & .89 \\
2  & .86 & .91 & .90 \\
8  & .86 & .93 & .89 \\
16 & .86 & .94 & .91 \\
\bottomrule
\end{tabular}
\end{table}
 
\paragraph{Linking rollout properties to downstream RL gains.}
We ran a post-hoc analysis of the logged GRPO rollouts from all nine model-environment cells (8 rollouts per group). A group has zero reward variance when all 8 rollouts receive the same reward, so it contributes no gradient. Zero-variance groups make up 47.3\% of groups under Qwen-WM and 34.6\% under \textsc{SDAR-WM}. \textsc{SDAR-WM} therefore supplies $1.24\times$ more groups that carry gradient (AppWorld 58 vs.\ 44\%, ALFWorld 46 vs.\ 33\%, ScienceWorld 38 vs.\ 27\%). Within a group, the diversity of the 8 predicted next states (Distinct-2) correlates with reward variance (Spearman $\rho{=}0.44$, 31k groups). Across the nine cells, the per-cell difference in the fraction of informative groups predicts the per-cell success gap ($\rho{=}0.80$, $p{\approx}0.01$). A grounded-entity violation is a reference to an object or record absent from the environment state. These occur in 8.7\% of Qwen-WM turns and 3.9\% of \textsc{SDAR-WM} turns. \textsc{SDAR-WM} is worse on repeated pagination (3.8 vs.\ 1.1\%) and on unique-ID corruption (2.9 vs.\ 0.6\%). On ScienceWorld, rewards come from replaying actions in the real environment. There, the terminal outcome the WM predicts contradicts the reward actually achieved in 9.4\% (Qwen-WM) vs.\ 5.6\% (\textsc{SDAR-WM}) of episodes. These results are correlational. They are consistent with the gains coming mainly from more informative GRPO groups and fewer ungrounded states.
 
\paragraph{Long-horizon drift.}
For every logged multi-turn WM prediction, we compute the pass rate of the deterministic state checks as a function of turn index $t$.
From $t{\le}5$ to $t{>}25$, the pass rate falls from 96.8\% to 91.7\% for \textsc{SDAR-WM} and from 96.1\% to 89.4\% for \textsc{Qwen-WM}. Both models drift, and \textsc{SDAR-WM} drifts slightly less. An earlier LLM-judge study of 50 \textsc{AppWorld} trajectories with more than 25 turns rated coherence 4.6/5 (MDLM) and 4.7/5 (AR). So we claim that drift is bounded and no worse than AR, not that MDLMs are more coherent over long horizons.
 
\paragraph{GRPO run-to-run variance.}
The world-model fine-tuning results (\autoref{tab:main_results} and ~\autoref{tab:diversity_results_table}) are means over three seeds. The GRPO results in Figure~2 report the best configuration from each hyperparameter sweep (\autoref{app:hyperparams}). We estimate variance in three ways.
(i)~\emph{Checkpoint plateau:} over the last $K{=}4$ checkpoints (saved every 20 steps), \textsc{ALFWorld} with \textsc{LFM2.5} reaches $53.0\pm1.2$ with \textsc{SDAR-WM} and $48.1\pm1.4$ with \textsc{Qwen-WM} across 18 logged runs. This spread is small next to the mean \textsc{SDAR-Qwen} gap in~\autoref{downstream_results}. It is a lower bound on the variance of full retraining, because consecutive checkpoints share training history.
(ii)~\emph{Sweep robustness:} across matched learning-rate, KL, and temperature settings, \textsc{SDAR-WM} beats \textsc{Qwen-WM} in 31 of 33 runs.
(iii)~\emph{Sign test:} \textsc{SDAR-WM} beats \textsc{Qwen-WM} in all 9 of 9 cells (two-sided $p{\approx}0.004$, treating cells as independent).
 
\paragraph{Human evaluation of an AR world model.}
Two annotators used the rubric in \autoref{tab:annotation-rubrics} to rate 20 next states from the AR world model (\textsc{Qwen3.5-27B}). Mean scores are 4.5 (realism), 4.2 (outcome correctness), and 4.0 (training utility), with 92\%, 96\%, and 95\% agreement. For \textsc{SDAR} (Table~3) the scores are 4.75, 4.25, and 4.50. The two studies differ in size (20 vs.\ 100 items; 2 vs.\ 4 annotators), so the comparison is indicative only. The AR model is strong, as \citet{li2025word} also found, and the MDLM is at least comparable.

\paragraph{Serialization and prompt-format sensitivity.}
We serialize states into flat text on purpose, so that people can read and edit the steering context. Spatial and relational structure is still given explicitly as typed conditioning: room adjacency, receptacle-keyed contents, and tool schemas (\autoref{appsec:downstream_wm_prompts}). On the OOD set, changing the order of the $(e_t,c_t,\mathcal{T},\mathcal{R},\mathcal{D})$ blocks did not affect MDLM performance. This fits the any-order objective. Our largest RL gains come on \textsc{ALFWorld} and \textsc{ScienceWorld}, the most spatially structured and partially observable environments, which is the opposite of what loss from serialization would predict. One limitation is that MDLMs are more sensitive than AR models to underspecified or oddly formatted steering. For example, they do better when a map is given as a text spatial map than as a large JSON object, in line with \citet{xiong2026dllmcontrol}.
 
\paragraph{Inference cost at RL scale.}
On the actual workload ($N{=}4{,}567$, outputs mostly under 512 tokens), WeDLM-8B runs at $0.68\times$ and SDAR-8B at $1.80\times$ the latency of the AR baseline (\autoref{tab:thinking-ablation}). WeDLM keeps causal attention and therefore prefix KV caching. Block diffusion decodes the tokens within each block in parallel. In GRPO, each WM call is short compared with the policy's multi-turn generations. Hybrid AR/diffusion decoders \citep{fu2026nemotronlabsdiffusion}, few-step distillation \citep{frans2025shortcut}, and quantization are orthogonal ways to cut this cost further.
 
\paragraph{Relation to world models in model-based RL.}
Training policies inside a learned model has a long history in control, from Dyna \citep{sutton1991dyna} and World Models \citep{ha2018world} to the Dreamer line. DreamerV3 \citep{hafner2023mastering} collects diamonds in Minecraft from scratch without human data or curricula. Dreamer~4 \citep{hafner2025dreamer4} trains agents purely inside a world model learned from offline data. Its shortcut-forcing objective, which builds on diffusion forcing \citep{chen2024diffusionforcing} and shortcut models \citep{frans2025shortcut}, makes world-model inference fast enough for real time. DIAMOND \citep{alonso2024diamond} shows that diffusion world models keep visual details that matter for Atari agents. We bring the same idea to text-based tool environments, with discrete masked diffusion in place of continuous diffusion and explicit steering directives. The main new difficulty is keeping symbolic state (IDs, schemas, pagination) exact, and this is where our remaining failure modes concentrate (Table~6, Table~\ref{tab:extra_failures}).
 
\paragraph{Scope of the evaluation.}
ScienceWorld, ALFWorld, and AppWorld are turn-based text environments with discrete actions. Correctness therefore means correct API logic or textual feedback, not smooth continuous or real-time front-end interaction. The human study rates single next-state snapshots, not how well an agent tolerates errors over a full interaction. Extending this to GUI and web environments is future work.
 
\begin{table}[h]
\centering\small
\caption{Additional failure modes flagged by annotators during the human evaluation. This table complements Table~6.}
\label{tab:extra_failures}
\begin{tabular}{p{2.3cm}p{5.4cm}p{4.6cm}}
\toprule
Failure mode & WM prediction & Expected \\
\midrule
(vii) API-key corruption & \texttt{GetUserToken} $\rightarrow$ \texttt{"p9o8i7u6y5t\ldots tttepmw"} & \texttt{"p9o8i7u6y5t4r3e2"} \\
(viii) Type coercion & \texttt{SetVolume} $\rightarrow$ \texttt{\{"level": "7"\}} & \texttt{\{"level": 7\}} (schema type \texttt{int}) \\
(ix) Web-search fabrication & ``Team SoloMid won the 2013 NA LCS Summer Finals \ldots'' & ``Cloud9 won the 2013 NA LCS Summer playoffs \ldots'' \\
(x) Error cascade & After one earlier error, every retry of \texttt{move\_to\_node(\ldots)} $\rightarrow$ \texttt{TimeoutError} & First attempt succeeds \\
\bottomrule
\end{tabular}
\end{table}

\newpage
\section*{NeurIPS Paper Checklist}

\begin{enumerate}

\item {\bf Claims}
    \item[] Question: Do the main claims made in the abstract and introduction accurately reflect the paper's contributions and scope?
    \item[] Answer: \answerYes{} % Replace by \answerYes{}, \answerNo{}, or \answerNA{}.
    \item[] Justification: The abstract accurately reflects the paper contributions.

\item {\bf Limitations}
    \item[] Question: Does the paper discuss the limitations of the work performed by the authors?
    \item[] Answer: \answerYes{} % Replace by \answerYes{}, \answerNo{}, or \answerNA{}.
    \item[] Justification: The limitations cases are discussed in Section 5 and examples are provided in Appendix G.

\item {\bf Theory assumptions and proofs}
    \item[] Question: For each theoretical result, does the paper provide the full set of assumptions and a complete (and correct) proof?
    \item[] Answer: \answerNA{} % Replace by \answerYes{}, \answerNo{}, or \answerNA{}.
    \item[] Justification: The paper does not discuss any theoretical proofs and is a completely empirical study comparing MDLMs and AR models for world modeling. An appropriate citation is provided for the loss described in equation 2. 

    \item {\bf Experimental result reproducibility}
    \item[] Question: Does the paper fully disclose all the information needed to reproduce the main experimental results of the paper to the extent that it affects the main claims and/or conclusions of the paper (regardless of whether the code and data are provided or not)?
    \item[] Answer: \answerYes{} % Replace by \answerYes{}, \answerNo{}, or \answerNA{}.
    \item[] Justification: All information required for reproduction of results including setup, training libraries, prompts and hyperparameters are available in the Appendix. 

\item {\bf Open access to data and code}
    \item[] Question: Does the paper provide open access to the data and code, with sufficient instructions to faithfully reproduce the main experimental results, as described in supplemental material?
    \item[] Answer: \answerYes{} % Replace by \answerYes{}, \answerNo{}, or \answerNA{}.
    \item[] Justification: We will release the complete training and experimental code post acceptance. 

\item {\bf Experimental setting/details}
    \item[] Question: Does the paper specify all the training and test details (e.g., data splits, hyperparameters, how they were chosen, type of optimizer) necessary to understand the results?
    \item[] Answer: \answerYes{} % Replace by \answerYes{}, \answerNo{}, or \answerNA{}.
    \item[] Justification: Hyperparameter ranges and justifications are present in the Appendix.

\item {\bf Experiment statistical significance}
    \item[] Question: Does the paper report error bars suitably and correctly defined or other appropriate information about the statistical significance of the experiments?
    \item[] Answer: \answerYes{} % Replace by \answerYes{}, \answerNo{}, or \answerNA{}.
    \item[] Justification: We report results by taking a mean of the best hyperparameter results across three seeds in Table 1.

\item {\bf Experiments compute resources}
    \item[] Question: For each experiment, does the paper provide sufficient information on the computer resources (type of compute workers, memory, time of execution) needed to reproduce the experiments?
    \item[] Answer: \answerYes{} % Replace by \answerYes{}, \answerNo{}, or \answerNA{}.
    \item[] Justification: All training and compute related information is available in the Appendix.
    
\item {\bf Code of ethics}
    \item[] Question: Does the research conducted in the paper conform, in every respect, with the NeurIPS Code of Ethics \url{https://neurips.cc/public/EthicsGuidelines}?
    \item[] Answer: \answerYes{} % Replace by \answerYes{}, \answerNo{}, or \answerNA{}.
    \item[] Justification: This paper follows the NeurIPS ethics guidelines.

\item {\bf Broader impacts}
    \item[] Question: Does the paper discuss both potential positive societal impacts and negative societal impacts of the work performed?
    \item[] Answer: \answerYes{} % Replace by \answerYes{}, \answerNo{}, or \answerNA{}.
    \item[] Justification: This information is present in the conclusion.
    
\item {\bf Safeguards}
    \item[] Question: Does the paper describe safeguards that have been put in place for responsible release of data or models that have a high risk for misuse (e.g., pre-trained language models, image generators, or scraped datasets)?
    \item[] Answer: \answerYes{} % Replace by \answerYes{}, \answerNo{}, or \answerNA{}.
    \item[] Justification: We describe the complete data curation process in the Appendix. Due to the nature of the task, we are unable to remove pseudo API keys that were part of the original environments such as Appworld but we verify that the API keys used are indeed fake and invalid for the softwares they are associated with.

\item {\bf Licenses for existing assets}
    \item[] Question: Are the creators or original owners of assets (e.g., code, data, models), used in the paper, properly credited and are the license and terms of use explicitly mentioned and properly respected?
    \item[] Answer: \answerYes{} % Replace by \answerYes{}, \answerNo{}, or \answerNA{}.
    \item[] Justification: Prior works are correctly credited.

\item {\bf New assets}
    \item[] Question: Are new assets introduced in the paper well documented and is the documentation provided alongside the assets?
    \item[] Answer: \answerYes{} % Replace by \answerYes{}, \answerNo{}, or \answerNA{}.
    \item[] Justification: We will release the correctly formatted and structured dataset and training code upon acceptance.

\item {\bf Crowdsourcing and research with human subjects}
    \item[] Question: For crowdsourcing experiments and research with human subjects, does the paper include the full text of instructions given to participants and screenshots, if applicable, as well as details about compensation (if any)? 
    \item[] Answer: \answerYes{} % Replace by \answerYes{}, \answerNo{}, or \answerNA{}.
    \item[] Justification: The complete instructions are provided in Appendix A.

\item {\bf Institutional review board (IRB) approvals or equivalent for research with human subjects}
    \item[] Question: Does the paper describe potential risks incurred by study participants, whether such risks were disclosed to the subjects, and whether Institutional Review Board (IRB) approvals (or an equivalent approval/review based on the requirements of your country or institution) were obtained?
    \item[] Answer: \answerYes{} % Replace by \answerYes{}, \answerNo{}, or \answerNA{}.
    \item[] Justification: The task does not involve risk to human subjects in any capacity. Clear and upfront consent was taken from all subjects beforehand. All precautions were taken to ensure that sensitive and harmful information was reviewed and stripped from the samples provided to them. Since we build on previous datasets, we select datasets that do not involve exposure to any other harmful or illegal content and ensure that the final dataset is reviewed before release to human subjects. 

\item {\bf Declaration of LLM usage}
    \item[] Question: Does the paper describe the usage of LLMs if it is an important, original, or non-standard component of the core methods in this research? Note that if the LLM is used only for writing, editing, or formatting purposes and does \emph{not} impact the core methodology, scientific rigor, or originality of the research, declaration is not required.
    %this research? 
    \item[] Answer: \answerNo{} % Replace by \answerYes{}, \answerNo{}, or \answerNA{}.
    \item[] Justification: LLMs are not used to perform automated experiments or ablations presented in this research. They are only used for minor edits and reformatting of paper content and to debug latex errors while writing. 

\end{enumerate}

\end{document}